\documentclass[11pt]{article}

\usepackage{acl}

\usepackage{times}
\usepackage{latexsym}

\usepackage[T1]{fontenc}

\usepackage[utf8]{inputenc}

\usepackage{microtype}

\usepackage{inconsolata}

\usepackage{graphicx}

\usepackage{subcaption}
\usepackage{booktabs}
\usepackage{tabularx,multirow,multicol}
\usepackage{array}
\usepackage{makecell}
\usepackage[table]{xcolor}
\usepackage[most]{tcolorbox}
\usepackage{listings}
\usepackage{csquotes}
\usepackage{amsmath,amssymb,mathtools,amsthm}
\usepackage{tikz}
\usepackage{float}      
\usepackage{placeins}   
\usepackage[capitalize,noabbrev]{cleveref}
\usepackage{enumitem}
\setlist{topsep=4pt,partopsep=0pt,parsep=0pt,itemsep=3pt,leftmargin=*}

\definecolor{boxframe}{RGB}{60,80,120}
\definecolor{boxback}{RGB}{246,248,252}
\newtcolorbox{promptbox}[1]{%
  enhanced, breakable,
  colback=boxback, colframe=boxframe,
  coltitle=white, fonttitle=\small\bfseries,
  colbacktitle=boxframe,
  boxrule=0.6pt, arc=1.5pt,
  left=6pt, right=6pt, top=5pt, bottom=5pt,
  boxsep=0pt,
  attach boxed title to top left={xshift=6pt, yshift=-2.4pt},
  boxed title style={boxrule=0pt, arc=1pt, left=4pt, right=4pt, top=1.5pt, bottom=1.5pt},
  title={#1},
  before skip=8pt, after skip=8pt,
  fontupper=\small}

\newcommand*\circled[1]{\tikz[baseline=(char.base)]{
            \node[shape=circle,draw,inner sep=1pt] (char) {#1};}}
\newcommand{\name}{UniData}
\newcommand{\namedataset}{UniDataset}

\title{UniData: Universal Multimodal Instruction Generation Pipeline}

\author{
\bfseries Jiaqi Tang\textsuperscript{1}\thanks{Equal contribution.},
Yi-Feng Wu\textsuperscript{2}\footnotemark[1],
Yuting Zhang\textsuperscript{3}\footnotemark[1],
Hao Lu\textsuperscript{3},
Bowen Fu\textsuperscript{4},
Qing-Guo Chen\textsuperscript{2}, \\
\bfseries Xiaogang Xu\textsuperscript{5},
Yuwei Hu\textsuperscript{2},
Shiyin Lu\textsuperscript{2},
Wei Wei\textsuperscript{4},
Lei Zhang\textsuperscript{4},
Zhao Xu\textsuperscript{2}, \\
\bfseries Weihua Luo\textsuperscript{2},
Qifeng Chen\textsuperscript{1}\thanks{Corresponding authors.},
Ying-Cong Chen\textsuperscript{3}\footnotemark[2] \\
\normalfont
\textsuperscript{1}The Hong Kong University of Science and Technology \quad
\textsuperscript{2}ATH, Alibaba Group \\
\textsuperscript{3}The Hong Kong University of Science and Technology (Guangzhou) \\
\textsuperscript{4}Northwestern Polytechnical University \quad
\textsuperscript{5}Zhejiang University \\
\texttt{\{jtang092, cqf\}@ust.hk},
\texttt{yingcongchen@hkust-gz.edu.cn}
}

\begin{document}
\maketitle

\begin{abstract}
Multimodal Large Language Models (MLLMs) are increasingly being applied in a wider range of real-world scenarios.
However, due to the substantial labor cost, creating high-quality multimodal instruction datasets for MLLMs remains a significant challenge.
Although some methods propose to generate instruction data, they often face limitations in modality support and struggle with generating multi-round instructions.
To address these problems, we introduce \name, a universal instruction generation pipeline, to transform simple user requirements into multi-round, multimodal instructions.
Specifically, \name\ first expands user requirements into multiple diverse events.
Using these events, \name\ then integrates an any-to-any large model for multimodal instruction generation.
Finally, \name\ enhances data quality by correcting irrelevant and redundant flows during inference, leveraging correlations between instruction rounds.
To train this pipeline, we also build \namedataset, a dataset comprising $20,000$ entries across nine modalities for improved multimodal generation. Our experiments demonstrate that \name\ achieves SOTA performance in data quality and can also enhance the understanding and generation capabilities of other multimodal models.

\end{abstract}

\section{Introduction}
\label{intro}

\begin{displayquote}
\itshape ``In God we trust; \\all others must bring data.''
\end{displayquote}
\begin{flushright}
\small\textemdash\ W. Edwards Deming
\end{flushright}

\begin{figure}[t]
    \centering
    \includegraphics[width=1\linewidth]{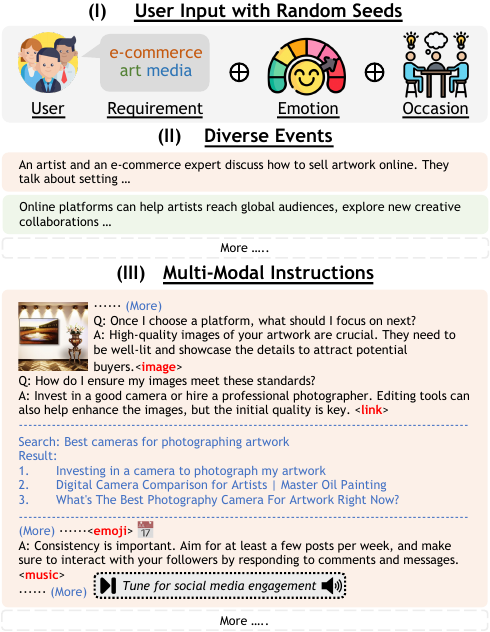}
    \caption{Usage of \name. \textbf{(I)} Users only input one simple requirement, combined with random emotions and occasions to create diverse events. \textbf{(II)} Each diverse event will generate one multimodal instruction. \textbf{(III)} The generated multimodal instructions include various modalities.}
    \label{motivation}
\end{figure}

Recently, Multimodal Large Language Models (MLLMs) have been extensively developed across various fields~\cite{cui2024survey,li2024embodied}.
However, to propel MLLMs toward fulfilling real-world requirements, data scarcity has gradually become a significant issue~\cite{liu2024visual, muller2017retrieval, yin2023lamm, wang2023vigc, wang2022self}.

To acquire high-quality data, most previous methods have relied on manual collection and annotation~\cite{li2017crowdsourced}, which consequently demands substantial human resources~\cite{tang2024hawk, liu2024mmbench, fang2024mmbenchv}.
While some approaches~\cite{li2023videochat, liu2024visual} have started utilizing GPT~\cite{openai2024gpt4ocard} for automated or semi-automated data generation to alleviate labor costs, most of them are designed for one specific scenario and cannot generalize to broader applications.

In this paper, we propose \name, a universal multimodal instruction generation pipeline. As shown in Figure~\ref{motivation}, by simply inputting basic requirements, users can finally obtain large-scale multimodal instruction datasets.

\begin{figure}[t]
    \centering
    \includegraphics[width=1\linewidth]{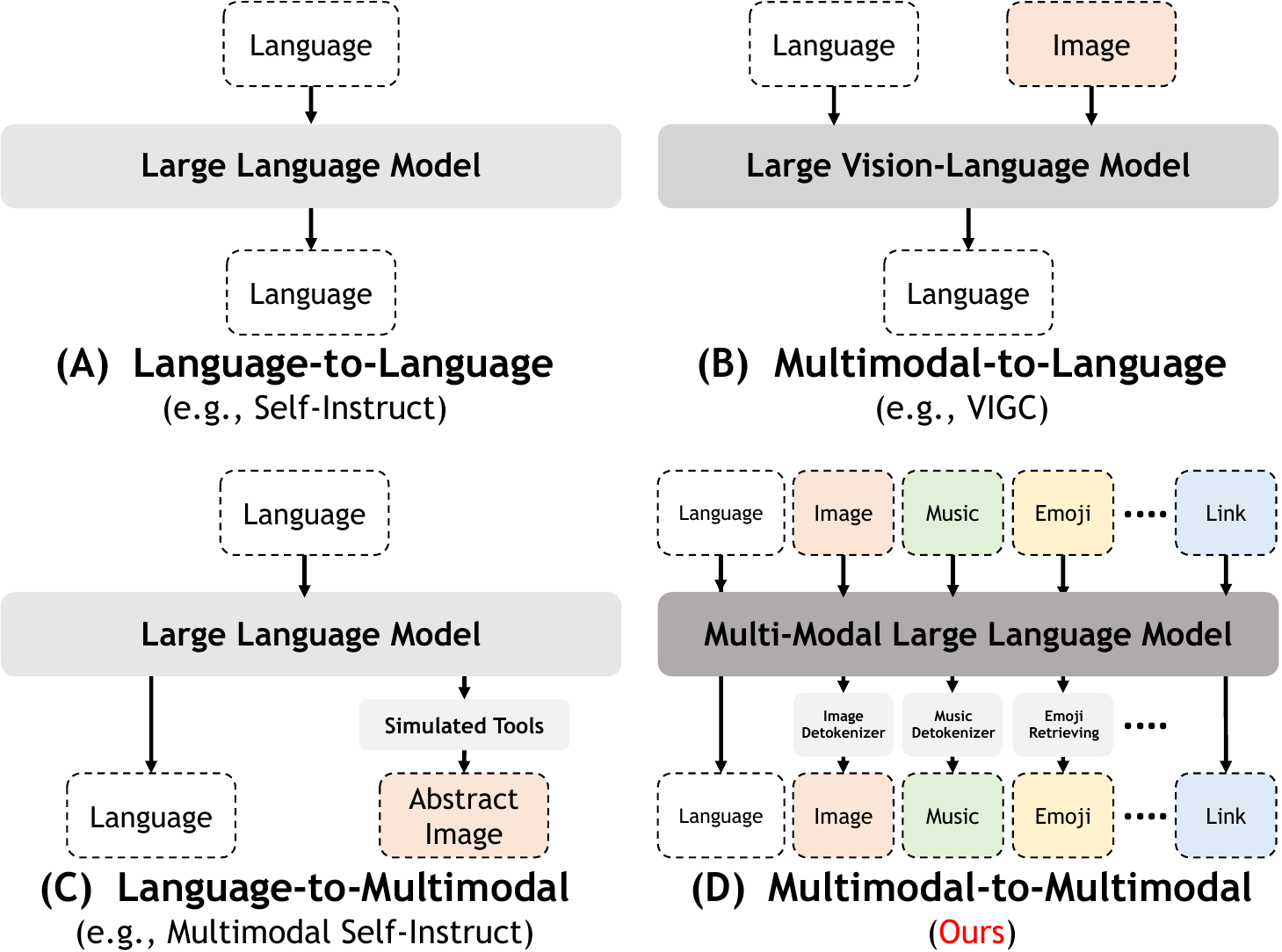}
    \caption{Comparison of different instruction generation frameworks. \textbf{(A)} Self-Instruct~\cite{wang2022self} can only support language modality. \textbf{(B)} VIGC~\cite{wang2023vigc} integrates visual modality into its generation. \textbf{(C)} Multimodal Self-Instruct~\cite{zhang2024multimodal} can output abstract images. \textbf{(D)} Ours (\name) can freely understand and generate different modalities in instructions.}
    \label{related}
\end{figure}

While some solutions have attempted to generate instructions in a universal setting, such as Self-Instruct~\cite{wang2022self} and VIGC~\cite{wang2023vigc}, these methods are primarily restricted to language modality (as shown in Figure~\ref{related} (A)(B)).
Multimodal Self-Instruct~\cite{zhang2024multimodal} attempts to generate instructions with corresponding abstract images, but its capabilities are limited to this (as shown in Figure~\ref{related} (C)). Furthermore, all of these methods provide limited support for multi-round instruction generation.

To address the aforementioned problems, we first focus on enhancing data diversity by expanding user keyword requirements into a series of potential instructional events, as described in Section~\ref{m1}. Next, to improve multimodal support, we propose an any-to-any large model that assists users in generating multi-round, multimodal instructions iteratively, as described in Section~\ref{m2}.
Finally, to ensure data quality in a multi-round context, we correct the instruction flow by assessing the correlation between instructions from previous and current rounds. This process effectively minimizes redundancy and irrelevance during inference (Section~\ref{m3}).

To address the limitations in quantity, quality, and copyright issues of existing data (in Section~\ref{datasection}), we also construct \namedataset\ to support the training of \name. This dataset comprises 20,000 entries and incorporates nine distinct modalities on both the input and output sides. These modalities include \texttt{<language>}, \texttt{<image>}, \texttt{<music>}, \texttt{<emoji>}, \texttt{<code>}, \texttt{<map>}, \texttt{<link>}, \texttt{<math>}, and \texttt{<QR\,\,code>} (see Section~\ref{datasection-procedure}). The inclusion of these diverse modalities is intended to enhance our data generation pipeline, enabling the creation of more comprehensive multimodal instructions and thereby improving overall versatility.

Experimental results demonstrate that the quality of our generated data surpasses that of other generation pipelines on all four dimensions we evaluate: reasonableness, clarity, detail, and relevance.
Furthermore, our generated data can also enhance the understanding and generation capabilities of other multimodal large language models.
Our contribution is shown as follows:
\begin{itemize}
\item We propose \name, a universal multimodal instruction generation pipeline that can create multi-round multimodal instructions from one simple requirement.
\item We propose \namedataset, a multimodal instruction dataset comprising $20,000$ instructions with nine distinct modalities in both input and output. \namedataset\ addresses existing limitations in data quantity, quality, and copyright status, and provides a robust data foundation for training \name\ or other any-to-any models.
\item Through comprehensive experiments, we demonstrate that \name\ surpasses existing pipelines in data quality, achieving SOTA performance in reasonableness, clarity, detail, and relevance. Additionally, the generated data can improve the understanding and generation capabilities of other multimodal models.
\end{itemize}

\section{Related Work}
\label{sec:related}

\paragraph{Multimodal Large Language Models}
Multimodal Large Language Models can understand multimodal information~\cite{survey1} by integrating various modalities, including but not limited to text, images, audio, and video.
Due to their versatility, they have been widely applied in diverse fields, such as embodied intelligence~\cite{li2024embodied}, video understanding~\cite{tang2024hawk}, and autonomous driving~\cite{cui2024survey}.

Most previous multimodal models~\cite{liu2024visual, li2023videochat, lu2024ovis} were designed to understand multimodal information, yet their output is limited to language only.
Recently, SEED-X~\cite{ge2024seed} and Show-o~\cite{xie2024showo} have begun to explore the integrated capabilities of image understanding and generation. Additionally, NExT-GPT~\cite{wu24next} has advanced by integrating more modalities, including images, audio, and video, for unified understanding and generation.

Nonetheless, training these models encounters substantial challenges, particularly regarding data availability. For example, NExT-GPT~\cite{wu24next} is unable to provide training data due to copyright constraints. Furthermore, Show-o~\cite{xie2024showo} is limited to training in an end-to-end way due to the lack of an appropriate dataset. Motivated by these challenges, we propose \name, which is designed to generate multimodal instructions to facilitate the model training process and potentially support further applications.

\paragraph{Instruction Generation}
The scarcity of training data impacts the performance of Multimodal Large Language Models, leading to increased interest in how to generate high-quality instruction data.
Self-Instruct~\cite{wang2022self} pioneered the generation of diverse instructions using a small set of seed tasks, but it focuses only on the language modality (Figure~\ref{related} (A)).
To overcome these limitations, VIGC~\cite{wang2023vigc} integrates visual understanding into the instruction generation process (Figure~\ref{related} (B)).
Besides, Multimodal Self-Instruct~\cite{zhang2024multimodal} addresses the problem of abstract image synthesis in instruction generation, utilizing synthetic tools to create both diagrams and corresponding text (Figure~\ref{related} (C)).
However, these methods still support limited multimodal capabilities.

To address these constraints, we propose \name, a universal multimodal instruction generation pipeline designed to both understand and generate nine types of multimodal information (Figure~\ref{related} (D)). This approach broadens the applicability of multimodal instruction generation compared to previous research.

\begin{figure*}[t]
    \centering
    \includegraphics[width=1\linewidth]{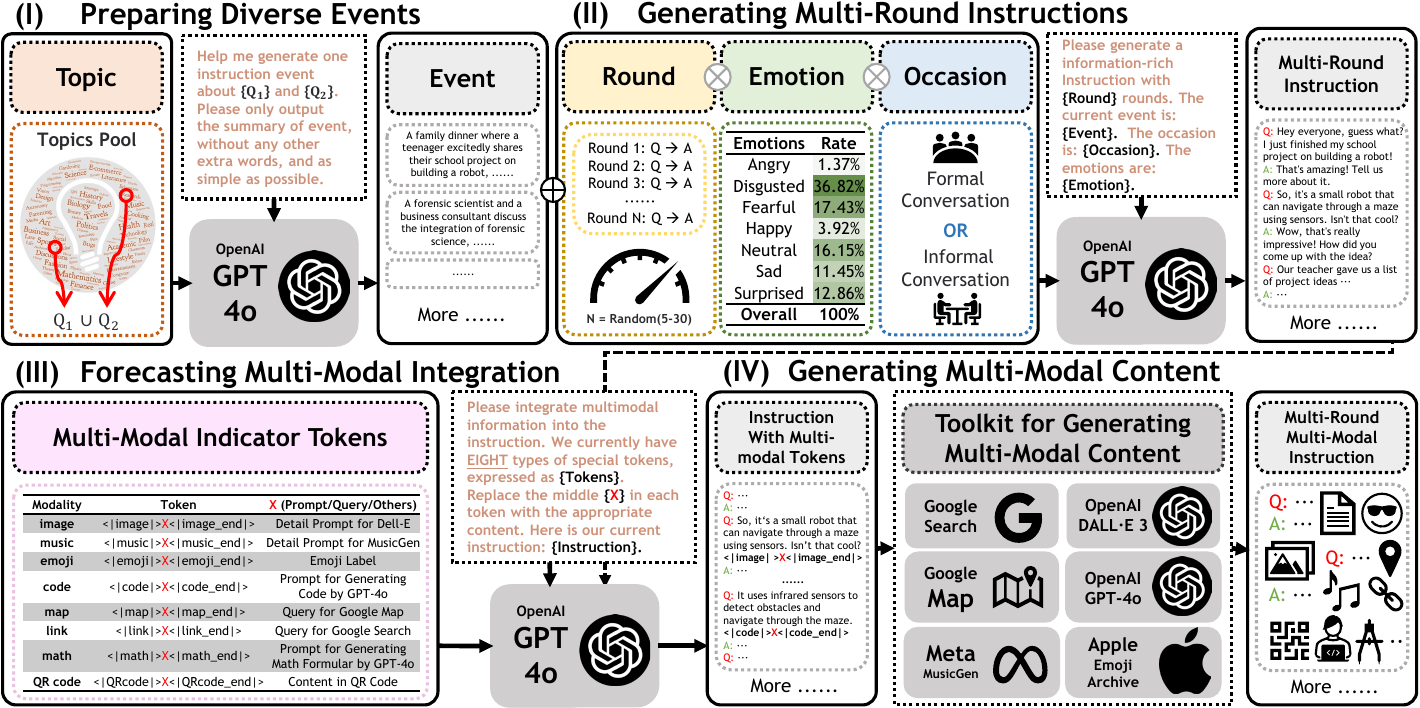}
    \caption{Procedure of data construction. \textbf{Stage (I)} prepares diverse events by random topic pairs. \textbf{Stage (II)} leverages these events to produce multi-round instructions. \textbf{Stage (III)} forecasts eight types of multimodal indicator tokens and integrates these into instructions. \textbf{Stage (IV)} uses a toolkit to generate multimodal content.}
    \label{dataset}
\end{figure*}

\section{Dataset Construction} \label{datasection}
To train \name, a dataset containing both multimodal inputs and outputs is essential. However, datasets meeting these criteria are scarce. As summarised in Table~\ref{datacompared} (Appendix~\ref{app:diversity}), the majority of existing datasets are confined to a single modality, especially for the output.
While NExT-GPT (MosIT)~\cite{wu24next} and GenHowTo~\cite{soucek2024genhowto} have developed some initial datasets encompassing multimodal inputs and outputs, they are hindered by a restricted range of modalities, limited instruction rounds, and a small data size. Furthermore, copyright concerns limit their availability.

To enrich our data resources, we have also developed a multimodal dataset named \namedataset\ (the \textbf{Ours} row in Table~\ref{datacompared}). This dataset supports inputs and outputs across nine modalities, thereby facilitating the training process of \name.

\subsection{Procedure of Data Construction} \label{datasection-procedure}
\paragraph{Stage (I): Preparing Diverse Events} 
To accommodate data diversity, it is crucial to generate varied \texttt{<events>} (tasks) for each multi-round instruction. Drawing inspiration from Self-Instruct~\cite{wang2022self}, we first create a \textit{Topic Pool} with $90$ \texttt{<topics>} spanning multiple domains, such as e-commerce, lifestyle, and sports. Then, we randomly pair two topics and input them into GPT-4o~\cite{openai2024gpt4ocard} to generate diverse \texttt{<events>}, as shown in Figure~\ref{dataset} (I).
Compared to a single \texttt{<topic>}, this approach introduces greater randomness ($\binom{90}{2}=4{,}005$ unique combinations), thereby facilitating data diversity.
\paragraph{Stage (II): Generating Multi-Round Instructions}
After obtaining \texttt{<events>} (tasks), we utilize them to generate multi-round instructions.
Considering other attributes of instructions, we also randomly choose different \texttt{<rounds>} (ranging from $5$ to $30$), \texttt{<emotion>} (random proportions of seven distinct emotions), and \texttt{<occasion>} (formal or informal) to collectively guide GPT-4o~\cite{openai2024gpt4ocard} in generating diverse instructions, as shown in Figure~\ref{dataset} (II).
Ultimately, we generate $20,000$ multi-round instructions.

\paragraph{Stage (III): Forecasting Multimodal Integration}
Our current multi-round instructions are restricted to a single modality (\texttt{<language>}). To generate multimodal content, we utilize GPT-4o~\cite{openai2024gpt4ocard} to predict potential multimodal content and incorporate the relevant indicators.

Beyond the base \texttt{<language>} modality, we forecast indicators for eight modalities: \texttt{<image>}, \texttt{<music>}, \texttt{<emoji>}, \texttt{<code>}, \texttt{<map>}, \texttt{<link>}, \texttt{<math>}, and \texttt{<QR code>}. Additionally, we generate prompts that describe the corresponding multimodal content, as shown in Figure~\ref{dataset} (III).
For instance, we generate prompts for GPT-4o~\cite{openai2024gpt4ocard} for modalities such as \texttt{<image>}, \texttt{<code>}, and \texttt{<math>}. For \texttt{<map>} and \texttt{<link>}, we develop search queries intended for use with Google Search and Google Maps.

\paragraph{Stage (IV): Generating Multimodal Content}
In the final stage, we develop an advanced Toolkit for Generating Multimodal Content. This toolkit includes GPT-4o~\cite{openai2024gpt4ocard}, Google Search, Google Maps, DALL·E 3, MusicGen~\cite{musicgen}, and the Apple Emoji Archive to assist in producing multimodal content. 
We leverage the generated prompts/content/queries to drive this toolkit, enabling the creation of multimodal content.
\subsection{Data Format and Statistics}

Each \namedataset\ sample is generated by mapping a contextual triple to a multimodal \texttt{\textbf{instruction}} that contains nine typed slots, one per modality (Eq.~\eqref{datastructure}). Slots are filled by dedicated generators (Appendix~\ref{app:toolkit}) conditioned on a modality-specific cue (a \texttt{<prompt>}, \texttt{<label>}, \texttt{<query>}, or \texttt{<content>} string), while the \texttt{<language>} slot carries the dialogue text itself.
\begin{equation}
\label{datastructure}
\small
\begin{aligned}
&\bigl(\texttt{<event>},\, \texttt{<emotion>},\, \texttt{<occasion>}\bigr) \\
&\quad\xrightarrow{\;\text{Generate}\;}
\texttt{\textbf{instruction}}
\left\{
\begin{array}{r@{\;\to\;}l}
                   & \texttt{<language>}  \\
\texttt{<prompt>}  & \texttt{<image>}     \\
\texttt{<prompt>}  & \texttt{<music>}     \\
\texttt{<label>}   & \texttt{<emoji>}     \\
\texttt{<prompt>}  & \texttt{<code>}      \\
\texttt{<query>}   & \texttt{<map>}       \\
\texttt{<query>}   & \texttt{<link>}      \\
\texttt{<prompt>}  & \texttt{<math>}      \\
\texttt{<content>} & \texttt{<QR\,code>}.
\end{array}
\right.
\end{aligned}
\end{equation}

\section{Methodology}
\label{method}

Previous works, such as Multimodal Self-Instruct~\cite{zhang2024multimodal} and VIGC~\cite{wang2023vigc}, typically support only limited multimodal generation or understanding capabilities, which do not satisfy users' practical needs.
In real-world applications, users often prefer to provide only one or a few simple requirements to generate large volumes of high-quality data.
\paragraph{Overview} To address the aforementioned issues, we propose \name, a pipeline for generating multimodal instructions.
\textbf{Initially}, based on user-provided requirements, we develop an event generator that expands on the limited inputs to create diverse events in Section~\ref{m1}.
\textbf{Next}, to generate multimodal instructions, we construct a large model for multimodal understanding and generation in Section~\ref{m2}.
\textbf{Finally}, to mitigate potential biases during long-term multi-round instruction generation, we correct the instruction flow to filter out unreasonable data in Section~\ref{m3}.

\subsection{Expanding Diverse Events} \label{m1}
Typically, users seek to generate one high-quality instruction dataset by providing only a few keyword requirements ($\text{K}$), such that ($\text{K} \rightarrow \sum_{n=1}^{N} \text{I}_n$). 
However, if we treat one $\text{K}$ as a single \texttt{<event>} ($\text{T}$), it becomes challenging to produce a diverse set of instructions.

\begin{figure}[t]
    \centering
    \includegraphics[width=1\linewidth]{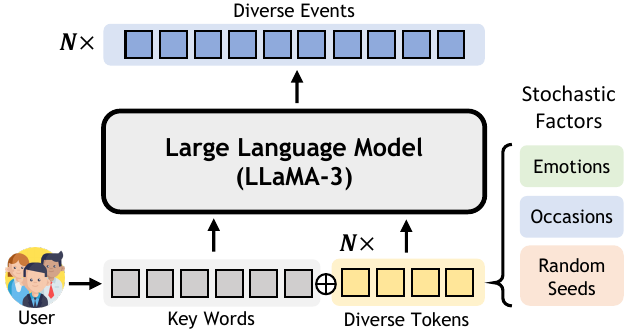}
    \caption{Pipeline of Diverse Event Generator ($\mathbf{DEG}$). Users can generate diverse events by inputting their needs.}
    \label{dgefigure}
\end{figure}

Therefore, we first need to expand the variety of $\text{T}$, so we build the Diverse Event Generator ($\mathbf{DEG}$) to expand diverse events, as shown in Figure~\ref{dgefigure} and Eq.~\eqref{deg}, as
\begin{equation} \label{deg} \small 
    \sum_{n=1}^{N} \text{I}_n \leftarrow \{ \text{T}_1, \text{T}_2, \ldots, \text{T}_N \} = \mathbf{DEG}(\text{K} \oplus \text{S}),
\end{equation} 
where $\oplus$ is the concatenation operator. The set $\{ \text{T}_1, \text{T}_2, \ldots, \text{T}_N \}$ denotes the collection of \texttt{<events>}, each corresponding to a distinct 
\texttt{instruction} ($\text{I}_n$). The term $\text{S}$ represents stochastic factors, such as random emotion proportions, random occasions, and other random seeds, which collectively enhance the capability of $\mathbf{DEG}(\cdot)$ to generate a broader spectrum of diverse events.
\subsection{Generating Multimodal Instructions} \label{m2}
\begin{figure*}[t]
    \centering
    \includegraphics[width=1\linewidth]{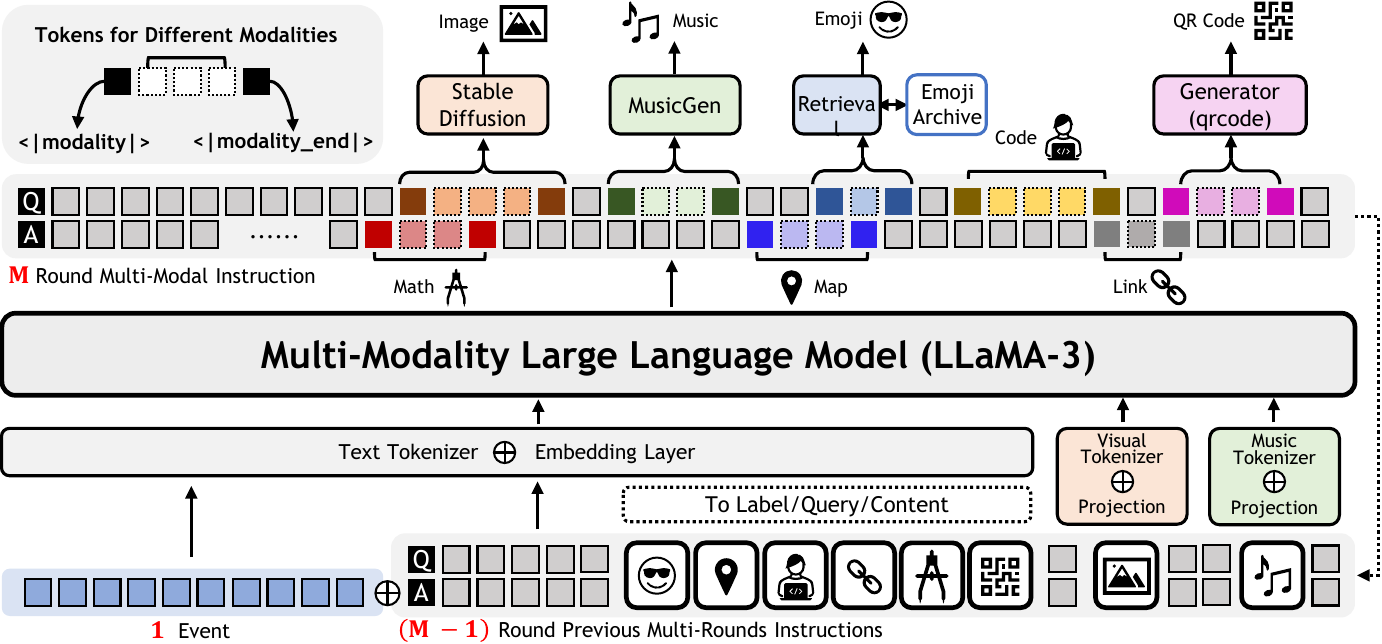}
    \caption{Pipeline of Multimodal Instruction Generator ($\mathbf{MIG}$). With just \textbf{one} event provided by the user, $\mathbf{MIG}$ can iteratively generate \textbf{multi-round multimodal} instructions based on the context.}
    \label{migfigure}
\end{figure*}

After obtaining diverse \texttt{<events>}, the next step is to generate multimodal instructions.
Inspired by previous any-to-any models~\cite{wu24next,xie2024showo}, we develop the Multimodal Instruction Generator ($\mathbf{MIG}$), a text-centred generator coupled with modality-specific realisation tools that can both understand and generate multimodal content, as shown in Figure~\ref{migfigure} and Eq.~\eqref{model},
\begin{equation} \label{model} \small
\text{I}_n^{M} = \mathbf{MIG}(\text{T}_{n} \oplus \sum_{m=1}^{M-1}\text{I}_n^m),
\end{equation} 
where $\mathbf{MIG}(\cdot)$ takes as input one event $\text{T}_{n}$ and previously generated multimodal instructions ($\sum_{m=1}^{M-1}\text{I}_n^m$), and predicts the next-round multimodal instruction ($\text{I}_n^{M}$).
\paragraph{Multimodal Tokenization} 
To generate the next-round instructions ($\text{I}_n^{M}$), $\mathbf{MIG}$ must understand the multimodal instruction from previous rounds ($\sum_{m=1}^{M-1}\text{I}_n^m$).
Thus, it is necessary to convert the previous multimodal instructions into token embeddings within a unified feature space.

Given that our training dataset, \namedataset, comprises nine distinct modalities, our tokenizing strategy ($\mathbf{F}(\cdot)$) is shown as Eq.~\eqref{tokenization},
\begin{equation} \label{tokenization} \small
    \mathbf{F}(\text{I}) = 
    \begin{bmatrix}
        \mathbf{f}_{\text{text}}[\text{I}_{\text{L}}, \mathbf{C}(\text{I}_{\text{E}} \oplus \text{I}_{\text{C}} \oplus \text{I}_{\text{MP}} \oplus \text{I}_{\text{LK}} \oplus \text{I}_{\text{MT}} \oplus \text{I}_{\text{Q}})] \\
        \mathbf{f}_{\text{img}}(\text{I}_{\text{I}}) \\
        \mathbf{f}_{\text{mus}}(\text{I}_{\text{M}})
    \end{bmatrix},
\end{equation}
where $\mathbf{f}_x(\cdot), x \in \{\text{text}, \text{img}, \text{mus}\}$ represent the tokenizer and embedding layer/projection for different modalities ($\text{I}_x, x \in \{ \text{L},\text{I},\text{M},\text{E},\text{C},\text{MP},\text{LK},\text{MT},\text{Q}\}$).

For modalities that are not inherently linguistic but can be represented as text ($\text{I}_x, x\in \{\text{E},\text{C},\text{MP},\text{LK},\text{MT},\text{Q}\}$), we can easily convert these into textual representations (prompt/label/query/content) by the formatter ($\mathbf{C}(\cdot)$). Then, these modalities can be combined with the language modality ($\text{I}_{\text{L}}$) and tokenized directly by using the default LLaMA Tokenizer~\cite{grattafiori2024llama3} ($\mathbf{f}_{\text{text}}(\cdot)$). The tokenized outputs are subsequently passed through a pre-trained embedding layer to generate token embeddings.

For \texttt{<image>} ($\text{I}_{\text{I}}$), we utilize the visual tokenizer and projection ($\mathbf{f}_{\text{img}}(\cdot)$) from Ovis-1.5~\cite{lu2024ovis} to convert images into visual embeddings.
For \texttt{<music>} ($\text{I}_{\text{M}}$), drawing on previous research~\cite{gong2023jointaudiospeechunderstanding, chu2023qwenaudioadvancinguniversalaudio, du2024lauragptlistenattendunderstand, verma2024whispergpthybridrepresentationaudio}, we integrate a pre-trained Whisper~\cite{radford2022robustspeechrecognitionlargescale} as the music tokenizer, followed by a Multi-Layer Perceptron (MLP) that projects into the desired embedding space.

\paragraph{Multimodal Indicator}
To distinguish different modalities, we introduce new special tokens to indicate the beginning and end of each modality.
As shown in Figure~\ref{migfigure}, the start of a modality is indicated by $\texttt{<|modality|>}$, where 
\begin{equation} \small
\begin{aligned}
    \texttt{modality} \in \{ \texttt{image}, \texttt{music},  \texttt{emoji}, \texttt{code},\\ \texttt{map}, \texttt{link}, \texttt{math}, \texttt{qrcode}\},
\end{aligned}
\end{equation}
and the end of the modality is indicated by $\texttt{<|modality\_end|>}$. This strategy allows the model to identify the position of multimodal insertion.

\paragraph{Architecture of Large Language Model} After aligning the multimodal signals into a unified embedding space, we select LLaMA-3~\cite{grattafiori2024llama3} as the main backbone of our pipeline. We also use the causal attention mechanism to sequentially predict the next token.

\paragraph{Multimodal Generation}
To generate multimodal instructions, we first need to establish multimodal token labels during training.
Although most modalities can be represented as text during tokenization, the image and music modalities lack corresponding text representations.
However, we find that during data construction (as described in Eq.~\eqref{datastructure}), we obtain $\texttt{<prompts>}$ ($\text{P}_{\text{I}}$ and $\text{P}_{\text{M}}$) for generating images and music.
Therefore, we can use these prompts as tokens to represent these two modalities discretely.
Eq.~\eqref{label} shows the ground truth label ($\mathbf{V}$),
\begin{equation}\small
\label{label}
\begin{aligned}
    \mathbf{V} = &\{\text{V}_{\text{L}}, (\text{V}_{\text{E}}, \text{V}_{\text{C}}, \text{V}_{\text{MP}}, \text{V}_{\text{LK}},\text{V}_{\text{MT}}, \text{V}_{\text{Q}}),\text{V}_{\text{I}},\text{V}_{\text{M}}\}, \ \  \\ &\text{where} \quad
    \text{V}_{\text{I}} = \mathbf{f}_{\text{text}}(\text{P}_{\text{I}}),
    \quad\text{V}_{\text{M}} = \mathbf{f}_{\text{text}}(\text{P}_{\text{M}}),
\end{aligned}
\end{equation}
and $\mathbf{f}_{\text{text}}(\cdot)$ is the tokenizer of LLaMA-3~\cite{grattafiori2024llama3} and $\text{V}_x, x \in \{\text{L},\text{I},\text{M},\text{E},\text{C},\text{MP},\text{LK},\text{MT},\text{Q}\}$ are the labels of different modalities. Our goal is to predict the next token ($\mathbf{V}^k$) given the previous tokens ($\{\mathbf{V}^i\}_{i=1}^{k-1}$) and the model parameters ($\Theta$), as Eq.~\eqref{loss},
\begin{equation} \label{loss} \small
\mathcal{L} = - \sum_{k} \log p(\mathbf{V}^k|\{\mathbf{V}^i\}_{i=1}^{k-1}; \Theta).
\end{equation}

\paragraph{Multimodal Detokenization} We use the default LLaMA-3 text detokenizer~\cite{grattafiori2024llama3} to convert all modalities into their text representations.
For \texttt{<code>}, \texttt{<map>}, \texttt{<link>}, and \texttt{<math>}, they can be directly detokenized into their original information.
For \texttt{<image>}, we use Stable Diffusion 3~\cite{sd3} to transform the text into the corresponding image.
For \texttt{<music>}, we employ MusicGen~\cite{musicgen} to convert the text into the corresponding music.
For \texttt{<emoji>}, we perform a similarity search of the output text to retrieve the corresponding emoji from the emoji archive.
For \texttt{<QR code>}, we can directly convert its content into a QR code by qrcode\footnote{\url{https://pypi.org/project/qrcode/}}.

\subsection{Correcting Instruction Flow} \label{m3}
Although now we can generate multimodal instructions by $\mathbf{MIG}$, during inference, it may also encounter common issues like repetitive generation or hallucinations~\cite{xu2022learning, Huang_2024}.
Therefore, cleaning and filtering data during long-term inference is crucial for maintaining data quality.
\begin{figure}[t]
    \centering
    \includegraphics[width=1\linewidth]{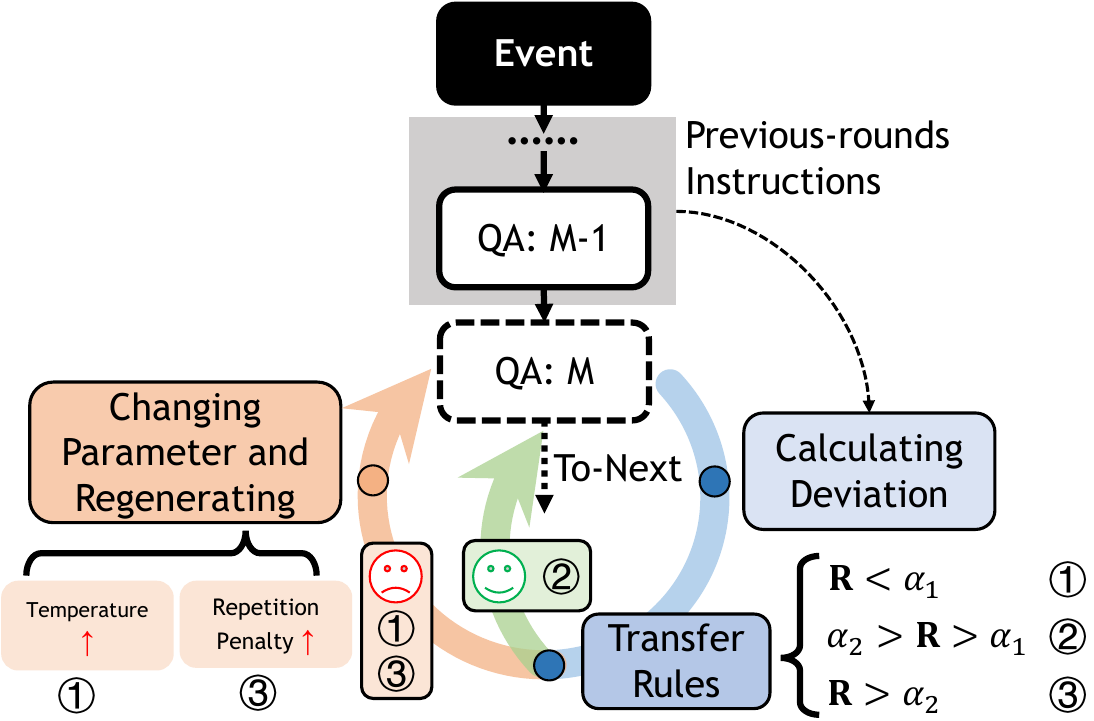}
    \caption{Inference chain for correcting instruction flow.}
    \label{inferchain}
\end{figure}
We design an inference chain in Figure~\ref{inferchain}.
We compare the newly generated one-round instruction ($\text{I}_n^M$) against the previous $k$-round instructions ($\sum_{m=M-k}^{M-1}\text{I}_n^m$) and apply the following three rules:
\begin{enumerate}
    \item \textbf{Repetitive Generation}: If the new instruction is close to the previous ones, increase the \textsc{Repetition Penalty} ($\uparrow$) to reduce repetition.
    \item \textbf{Irrelevant Generation}:  If the new instruction is unrelated to the previous ones, raise the \textsc{Temperature} ($\uparrow$) to introduce randomness, thereby sampling more alternatives.
    \item \textbf{High-Quality Generation}: If the instruction is neither repetitive nor irrelevant, we regard it as high-quality and proceed to the generation of the next round.
\end{enumerate}
To assess deviation, we use cosine similarity, as Eq.~\eqref{cosine},
\begin{equation} \label{cosine} \small
\mathbf{R} = \frac{1}{k}\sum_{m=M-k}^{M-1} \texttt{cos\_s}(\text{I}_n^M, \text{I}_n^m) = \frac{1}{k}\sum_{m=M-k}^{M-1}  \frac{\text{I}_n^M \cdot\text{I}_n^m}{|\text{I}_n^M| \left|\text{I}_n^m\right|}.
\end{equation}
If $\mathbf{R} < \alpha_1$, it indicates ``Irrelevant Generation''. If $\mathbf{R} > \alpha_2$, it indicates ``Repetitive Generation''. If $ \alpha_2 > \mathbf{R} > \alpha_1$, it indicates ``High-Quality Generation''. Empirically, we currently set $\alpha_1 = 0.05$ and $\alpha_2 = 0.4$ to minimize errors. Users can also adjust these values dynamically to maintain instruction quality and relevance based on their needs.

\begin{table*}[t]
  \centering
  \Large
  \caption{Quantitative performance in data quality. \textcolor{red}{Red} indicates the best performance.\small{ \textcolor{gray}{(\#Others includes all other modalities (\#Emoji, \#Code, \#Math, \#Map, \#Link, \#QR Code)).}}}
  \resizebox{1.0\linewidth}{!}{
    \begin{tabular}{l|l|cccc|ccc|c}
    \toprule
    \multirow{2}[2]{*}{\textbf{Method}} & \multirow{2}[2]{*}{Backbones} & \multicolumn{4}{c|}{GPT-Guided Metrics $(\mathbf{\uparrow})$~\cite{openai2024gpt4ocard}} & \multicolumn{3}{c|}{Avg. \#Multimodalities$(\mathbf{\uparrow})$} & \multirow{2}[2]{*}{Avg. \#Rounds} \\
    \cmidrule{3-9} 
    & & Reasonableness & Clarity & Detail & Relevance & \#Image & \#Music & \#Others & \\
    \midrule
    \textbf{Self-Instruct}~\cite{wang2022self} & GPT-4~\cite{openai2024gpt4ocard} & 0.624 & 0.444 & 0.594 & 0.520 & - & - & - & $\approx$ 3 \\
    \textbf{VIGC}~\cite{wang2023vigc} & Vicuna 7B~\cite{chiang2023vicuna} & 0.460 & 0.359 & 0.534 & 0.569 & - & - & - & $\approx$ 1 \\
    \midrule
    \textbf{Ours (\name)} & LLaMA-3 7B~\cite{grattafiori2024llama3} & \textcolor{red}{0.661} & \textcolor{red}{0.527} & \textcolor{red}{0.667} & \textcolor{red}{0.682} & \textcolor{red}{1.92} & \textcolor{red}{0.74} & \textcolor{red}{7.37} & $\approx$ \textcolor{red}{17.5}\\
    \bottomrule
    \end{tabular}}
  \label{qualtable}%
\end{table*}%
\section{Experiments}
\label{data}
In this section, we compare our instruction generation pipeline with others to evaluate instruction quality (see Section~\ref{e1}). Besides, we assess whether the generated instructions can improve multimodal generation and understanding performance after fine-tuning (see Section~\ref{e2}). Finally, we demonstrate the effectiveness of expanding events and correcting instruction flow (see Section~\ref{e3}).

Furthermore, we conduct a user study to assess the quality of the generated instructions, demonstrate the adaptability and balance of multimodal content to support different needs, validate performance on out-of-domain data, and provide additional qualitative results in the Appendix.

\paragraph{Training \& Testing}
During the pre-training phase, we adhere to the training strategy and dataset outlined in Ovis-1.5~\cite{lu2024ovis}, enabling the model to acquire initial multimodal understanding ability.
In the fine-tuning phase, we keep the parameters of the visual and music tokenizers fixed and randomly select $90\%$ of \namedataset\ for model fine-tuning, reserving the remaining $10\%$ for testing.

\subsection{Instruction Quality Evaluation} \label{e1}
\paragraph{Baselines} We select two representative instruction generation frameworks, Self-Instruct~\cite{wang2022self} and VIGC~\cite{wang2023vigc}, for comparison on our test data. In Self-Instruct~\cite{wang2022self}, due to the unavailability of GPT-3, we replace it with the latest GPT-4~\cite{openai2024gpt4ocard}. A discussion of more recent instruction-data generation methods and how they compare in scope is provided in Appendix~\ref{app:recent-baselines}.
\paragraph{Evaluation Metrics}
Following previous work~\cite{li2023videochat,tang2024hawk}, we use GPT-Guided metrics to evaluate the quality differences between generated results and references. Our goal is not to precisely match the ground truth but to produce high-quality instructions. We focus on four key metrics:
\circled{\textbf{1}} \textbf{Reasonableness} -  assesses if the instruction is logical and appropriate;
\circled{\textbf{2}} \textbf{Clarity} - ensures the instructions are clear and unambiguous;
\circled{\textbf{3}} \textbf{Detail} - evaluates whether the instructions include more detailed information;
\circled{\textbf{4}} \textbf{Relevance} - checks that the content is related to the desired outcome.

\begin{figure}[t]
    \centering
    \includegraphics[width=1\linewidth]{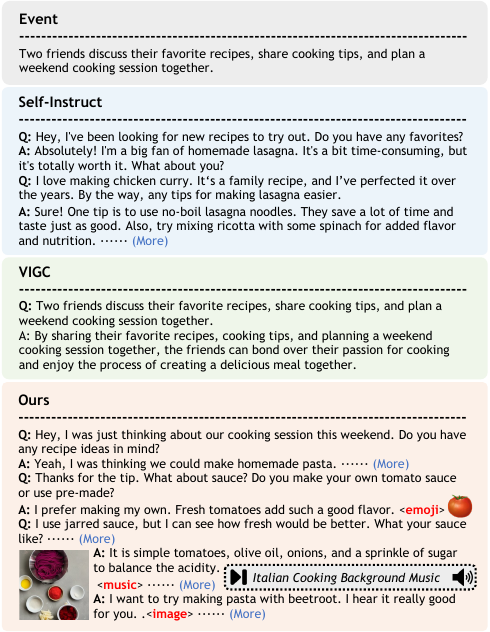}
    \caption{Qualitative comparison in data quality.}
    \label{qual}
\end{figure}
\paragraph{Performance Analysis} Table~\ref{qualtable} presents the quantitative results of our data quality. \name\ surpasses both baselines on all four dimensions, with the largest margin on Clarity and Relevance. Furthermore, \name\ supports more rounds of instruction generation and can produce rich multimodal content within instructions.
Figure~\ref{qual} shows the qualitative performance in data quality. Compared to other baselines that are limited to generating textual content, \name\ can generate high-quality multimodal instructions.

\subsection{Post Fine-Tuning Assessment}\label{e2}
Referring to previous research~\cite{wang2023vigc,wang2022self}, we also evaluate \textbf{the effectiveness of the generated multimodal instructions} (+ $\mathcal{I}_\texttt{\name} $) by fine-tuning other multimodal models.

\begin{table*}[t]
    \centering
    \Large
    \begin{minipage}{0.62\linewidth}
        \centering
        \caption{Improvement of music understanding.}
  \resizebox{\linewidth}{!}{
        \begin{tabular}{l|ccccc}
            \toprule
             & BLEU ($\uparrow$) & BLEU-4 ($\uparrow$) & METEOR ($\uparrow$) & ROUGE ($\uparrow$) & BERT ($\uparrow$)\\
            \midrule
            \textbf{Baseline} & 0.2675 & 0.2019 & 0.3232 & 0.3376 & 0.8932\\
            \textbf{Baseline} + $ \mathcal{I}_\texttt{\name}$ & \textcolor{red}{0.2681} & \textcolor{red}{0.2076} & \textcolor{red}{0.3330} & \textcolor{red}{0.3496} & \textcolor{red}{0.8936}\\
            \bottomrule
        \end{tabular}}
        \label{music_understanding}
    \end{minipage}%
    \begin{minipage}{0.37\linewidth}
        \centering
        \caption{Music generation.}
          \resizebox{\linewidth}{!}{
        \begin{tabular}{l|ccc}
            \toprule
             & FAD ($\downarrow$) & CLAP ($\uparrow$) & KL ($\downarrow$) \\
            \midrule
            \textbf{Baseline} & 16.69 & 0.3138 & 2.1358\\
            \textbf{Baseline} + $\mathcal{I}_\texttt{\name} $ & \textcolor{red}{16.51} & \textcolor{red}{0.3192} & \textcolor{red}{2.1038}\\
            \bottomrule
        \end{tabular}}
        \label{music_generation}
    \end{minipage}
      \begin{minipage}{0.5\textwidth}
        \centering
        \caption{Improvement on different modalities.}
        \resizebox{\linewidth}{!}{
            \begin{tabular}{l|cccc}
                \toprule
                \textbf{Modality} & Code $(\mathbf{\uparrow})$ & Math $(\mathbf{\uparrow})$ & Text-based $(\mathbf{\uparrow})$ & Reasoning $(\mathbf{\uparrow})$ \\
                \midrule
                \textbf{Benchmark} & MBPP & MathVista & MMLU-Pro & MathVision \\ 
                \midrule
                \textbf{Baseline} & 20.8 & 59.4 & 20.3 & 12.8 \\ 
                \textbf{Baseline + $\mathcal{I}_\texttt{\name} $} & \textcolor{red}{21.3} & \textcolor{red}{60.1} & \textcolor{red}{21.9} & \textcolor{red}{13.0} \\
                \bottomrule
            \end{tabular}
        }
        \label{otherm}
    \end{minipage}
    \begin{minipage}{0.48\textwidth}
    \caption{Image understanding/generation.}
    \resizebox{\linewidth}{!}{
        \begin{tabular}{l|ccc|c}
            \toprule
            \multirow{2}[4]{*}{} & \multicolumn{3}{c|}{Image Understanding} & \multirow{2}[4]{*}{\makecell{Image\\Generation} ($\uparrow$)} \\
            \cmidrule{2-4} & Single ($\uparrow$) & Multi ($\uparrow$) & Interleaved ($\uparrow$) &  \\
            \midrule
            \textbf{Baseline} & 26.19\% & 37.73\% & 19.42\% & 24.43\% \\
            \textbf{Baseline + $\mathcal{I}_\texttt{\name} $} & \textcolor{red}{27.15\%} & \textcolor{red}{39.09\%} & \textcolor{red}{35.25\%} & \textcolor{red}{25.90\%} \\
            \bottomrule
        \end{tabular}
    }
    \label{tab:addlabel}
    \end{minipage}
    \begin{minipage}{0.46\textwidth}
    \centering
    \caption{Improvement on visual understanding.}
    \Huge
    \resizebox{\linewidth}{!}{
    \begin{tabular}{l|ccc}
        \toprule
        \textbf{Benchmark} & MMMU $(\uparrow)$ & ChartQA $(\uparrow)$ & TextVQA $(\uparrow)$ \\
        \midrule
        \textbf{Baseline} & 35.8 & 17.4 & 58.9 \\ 
        \textbf{Baseline + $\mathcal{I}_\texttt{\name} $} & \textcolor{red}{36.4} & \textcolor{red}{17.9} & \textcolor{red}{60.1} \\
        \bottomrule
    \end{tabular}}
    \label{vunderstanding}
    \end{minipage}
    \begin{minipage}{0.52\textwidth}
        \centering
        \caption{Quantitative performance in ablation study.}
        \resizebox{\linewidth}{!}{
            \begin{tabular}{l|cccc|c}
                \toprule
                & Reasonableness $(\mathbf{\uparrow})$ & Clarity $(\mathbf{\uparrow})$ & Detail $(\mathbf{\uparrow})$ & Relevance $(\mathbf{\uparrow})$ & \#Ins. \\
                \midrule
                \textbf{w/o Expansion} & 0.557 & 0.518 & 0.607 & 0.550 & $k$\\
                \textbf{w/o Correction} & 0.648 & 0.489 & 0.610 & 0.654 & \textcolor{red}{$N \times k$}\\
                \textbf{Ours} & \textcolor{red}{0.661} & \textcolor{red}{0.527} & \textcolor{red}{0.667} & \textcolor{red}{0.682} & \textcolor{red}{$N \times k$} \\
                \bottomrule
            \end{tabular}
        }
        \label{ab2t}
    \end{minipage}
\end{table*}

\paragraph{Visual Understanding} We utilize a suite of benchmarks (MMMU~\cite{yue2024mmmu}, ChartQA~\cite{masry2022chartqa}, and TextVQA~\cite{singh2019towards}) to evaluate the effectiveness of our generated data in visual reasoning tasks. As detailed in Table~\ref{vunderstanding}, the integration of our data improves the performance of Ovis2-1B~\cite{lu2024ovis} in these benchmarks.

\paragraph{Image Understanding and Generation} We evaluate image understanding and generation capabilities using the SEED-X model~\cite{ge2024seed} and the SEED-Bench-2 benchmark~\cite{li2023seedbench2}. Our data improves both Image-to-Text and Text-to-Image tasks. As shown in Table~\ref{tab:addlabel}, the gain is largest on the interleaved setting ($19.42\%\!\rightarrow\!35.25\%$), with smaller gains on the single- and multi-image settings. Additionally, in the domain of image generation, our dataset helps the model produce relevant images based on textual inputs.
\paragraph{Music Understanding and Generation} We utilize the MuMu-LLaMA model~\cite{liu2024mumu} to assess advancements in music understanding and generation. As illustrated in Tables~\ref{music_understanding} and~\ref{music_generation}, our generated instructions improve both domains, although the margins are small. In the realm of music understanding (Table~\ref{music_understanding}), all five metrics move in the expected direction, with the clearest gains on METEOR and ROUGE. In music generation (Table~\ref{music_generation}), the fine-tuned model exhibits decreased FAD and KL scores alongside an elevated CLAP score, reflecting more coherent and high-quality music synthesis from textual prompts.
\paragraph{Code Generation} We evaluate the impact of our generated data on code modality by enhancing the code generation capabilities of a large language model, LLaMA 2~\cite{touvron2023llama2}, using the Mostly Basic Python Programming (MBPP) Benchmark~\cite{austin2021program}. As presented in Table~\ref{otherm}, our instructions improve performance.
\paragraph{Mathematical Problem} To evaluate the efficacy of our generated instructions in enhancing complex reasoning and mathematical problem-solving capabilities, we benchmark the Ovis2-1B~\cite{lu2024ovis} on the MathVista~\cite{lu2023mathvista} and MathVision~\cite{wang2024mathvision} datasets. In Table~\ref{otherm}, our instructions improve performance on both mathematical benchmarks. The gains are consistent in direction across the two, though modest in magnitude.
\paragraph{Other Text-based Modalities} To confirm the versatility of our data across various text-based modalities, including language, QR code, link, and map, we assessed their performance using a general language-based benchmark, MMLU-Pro~\cite{wang2024mmlu}. As shown in Table~\ref{otherm}, adding our data raises MMLU-Pro accuracy from $20.3$ to $21.9$, indicating that the text-based modalities also transfer to a general language benchmark.

\begin{figure}[t]
        \centering
        \includegraphics[width=\linewidth]{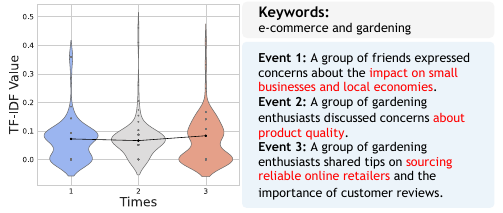}
        \caption{(\textbf{left}) Difference distribution of the events in three random generations. (\textbf{right}) Example of diverse events based on the same keywords.}
        \label{abp1}
\end{figure}

\subsection{Ablation Study} \label{e3}
\paragraph{Effectiveness of Expanding Diverse Events}
To demonstrate the effectiveness of expanding events, we randomly expand three events for each keyword on the test data to observe whether these events differ from one another (diversity). As shown in Figure~\ref{abp1} (\textbf{left}), the TF-IDF~\cite{ramos2003tfidf} feature distributions across three generation runs (on a random subset) show considerable diversity. In Figure~\ref{abp1} (\textbf{right}), we also provide an example showing that different events exhibit distinct content (\textcolor{red}{in highlighted text}).
In the \textbf{w/o Expansion} setting of Table~\ref{ab2t}, the generation of instructions relies solely on the few keywords provided by the user. Thus, the generation pipeline lacks the necessary enriched prompt information, which leads to a decline in the quality of generated instructions. Additionally, the number of instructions that can be generated is substantially reduced.

\begin{figure}[t]
        \centering
        \includegraphics[width=\linewidth]{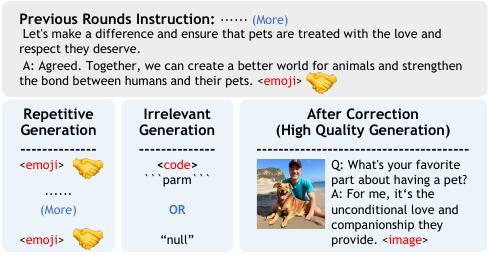}
        \caption{Example of flow error correction.}
        \label{ab2i}
\end{figure}
\paragraph{Effectiveness of Correcting Instruction Flow}
To demonstrate the effectiveness of correcting instruction flow, we run inference twice, \textit{with} and \textit{without} the correction, as shown in Table~\ref{ab2t} and Figure~\ref{ab2i}. 
Table~\ref{ab2t} shows that after correction, our data quality improves on all four dimensions.
Figure~\ref{ab2i} provides an example of the repetitive generation problem (same \texttt{<emojis>}) and irrelevant generation problem (\texttt{<code>} or ``null'') in multimodal content.
After correction, instruction generation can effectively avoid these errors.

\section{Conclusion}
This paper presents \name, a universal pipeline for multimodal instruction generation. As a fundamental contribution, \name\ establishes a framework for scalable data acquisition critical to MLLMs. By delivering high-quality, diverse, and cost-effective instruction sets, \name\ accelerates progress in multimodal learning paradigms and their downstream applications.

\section*{Limitations}

\name\ is a text-centered orchestration pipeline rather than a native any-to-any model, so per-modality quality is bounded by the external tools that realize each modality. Long-horizon structured reasoning is handled only by simple correction rules, and the nine-modality coverage still omits video, 3D, motion, tables, biological data, and spoken input. Finally, \name\ inherits biases from its LLM backbones, search APIs, and modality generators; these may enter \namedataset\ and be amplified downstream, and systematic bias audits remain future work.

\bibliography{custom}

@article{lu2024ovis,
  title={Ovis: Structural Embedding Alignment for Multimodal Large Language Model}, 
  author={Shiyin Lu and Yang Li and Qing-Guo Chen and Zhao Xu and Weihua Luo and Kaifu Zhang and Han-Jia Ye},
  year={2024},
  journal={arXiv:2405.20797}
}

@inproceedings{singh2019towards,
    title={Towards {VQA} Models That Can Read},
    author={Singh, Amanpreet and Natarajan, Vivek and Shah, Meet and Jiang, Yu and Chen, Xinlei and Batra, Dhruv and Parikh, Devi and Rohrbach, Marcus},
    booktitle={Proceedings of the IEEE/CVF Conference on Computer Vision and Pattern Recognition},
    pages={8317--8326},
    year={2019}
}

@inproceedings{masry2022chartqa,
  title={{ChartQA}: A Benchmark for Question Answering about Charts with Visual and Logical Reasoning},
  author={Masry, Ahmed and Long, Do Xuan and Tan, Jia Qing and Joty, Shafiq and Hoque, Enamul},
  booktitle={Findings of the Association for Computational Linguistics: ACL 2022},
  pages={2263--2279},
  year={2022},
  publisher={Association for Computational Linguistics}
}

@inproceedings{yue2024mmmu,
  title={Mmmu: A massive multi-discipline multimodal understanding and reasoning benchmark for expert agi},
  author={Yue, Xiang and Ni, Yuansheng and Zhang, Kai and Zheng, Tianyu and Liu, Ruoqi and Zhang, Ge and Stevens, Samuel and Jiang, Dongfu and Ren, Weiming and Sun, Yuxuan and others},
  booktitle={Proceedings of the IEEE/CVF Conference on Computer Vision and Pattern Recognition},
  pages={9556--9567},
  year={2024}
}

@inproceedings{wang2024mmlu,
  title={Mmlu-pro: A more robust and challenging multi-task language understanding benchmark},
  author={Wang, Yubo and Ma, Xueguang and Zhang, Ge and Ni, Yuansheng and Chandra, Abhranil and Guo, Shiguang and Ren, Weiming and Arulraj, Aaran and He, Xuan and Jiang, Ziyan and others},
  booktitle={The Thirty-eight Conference on Neural Information Processing Systems Datasets and Benchmarks Track},
  year={2024}
}

@inproceedings{wang2024mathvision,
  title={Measuring Multimodal Mathematical Reasoning with {MATH-Vision} Dataset},
  author={Wang, Ke and Pan, Junting and Shi, Weikang and Lu, Zimu and Zhan, Mingjie and Li, Hongsheng},
  booktitle={Advances in Neural Information Processing Systems (NeurIPS) Datasets and Benchmarks Track},
  year={2024}
}

@inproceedings{lu2023mathvista,
  title={{MathVista}: Evaluating Mathematical Reasoning of Foundation Models in Visual Contexts},
  author={Lu, Pan and Bansal, Hritik and Xia, Tony and Liu, Jiacheng and Li, Chunyuan and Hajishirzi, Hannaneh and Cheng, Hao and Chang, Kai-Wei and Galley, Michel and Gao, Jianfeng},
  booktitle={International Conference on Learning Representations (ICLR)},
  year={2024}
}

@article{austin2021program,
  title={Program synthesis with large language models},
  author={Austin, Jacob and Odena, Augustus and Nye, Maxwell and Bosma, Maarten and Michalewski, Henryk and Dohan, David and Jiang, Ellen and Cai, Carrie and Terry, Michael and Le, Quoc and others},
  journal={arXiv preprint arXiv:2108.07732},
  year={2021}
}

@inproceedings{ramos2003tfidf,
  title={Using tf-idf to determine word relevance in document queries},
  author={Ramos, Juan and others},
  booktitle={Proceedings of the first instructional conference on machine learning},
  volume={242},
  number={1},
  pages={29--48},
  year={2003},
  organization={Citeseer}
}

@article{li2023seedbench2,
  title={SEED-Bench-2: Benchmarking Multimodal Large Language Models},
  author={Li, Bohao and Ge, Yuying and Ge, Yixiao and Wang, Guangzhi and Wang, Rui and Zhang, Ruimao and Shan, Ying},
  journal={arXiv preprint arXiv:2311.17092},
  year={2023}
  }

@misc{chiang2023vicuna,
      title={Vicuna: An Open-Source Chatbot Impressing {GPT-4} with 90\%* {ChatGPT} Quality},
      author={Chiang, Wei-Lin and Li, Zhuohan and Lin, Zi and Sheng, Ying and Wu, Zhanghao and Zhang, Hao and Zheng, Lianmin and Zhuang, Siyuan and Zhuang, Yonghao and Gonzalez, Joseph E. and Stoica, Ion and Xing, Eric P.},
      year={2023},
      month={March},
      howpublished={\url{https://lmsys.org/blog/2023-03-30-vicuna/}}
}

@article{liu2024mumu,
  title={MuMu-LLaMA: Multi-modal Music Understanding and Generation via Large Language Models},
  author={Liu, Shansong and Hussain, Atin Sakkeer and Wu, Qilong and Sun, Chenshuo and Shan, Ying},
  journal={arXiv preprint arXiv:2412.06660},
  year={2024}
}

@inproceedings{fang2024mmbenchv,
  title={{MMBench-Video}: A Long-Form Multi-Shot Benchmark for Holistic Video Understanding},
  author={Fang, Xinyu and Mao, Kangrui and Duan, Haodong and Zhao, Xiangyu and Li, Yining and Lin, Dahua and Chen, Kai},
  booktitle={Advances in Neural Information Processing Systems (NeurIPS) Datasets and Benchmarks Track},
  year={2024}
}

@inproceedings{liu2024mmbench,
  title={{MMBench}: Is Your Multi-modal Model an All-around Player?},
  author={Liu, Yuan and Duan, Haodong and Zhang, Yuanhan and Li, Bo and Zhang, Songyang and Zhao, Wangbo and Yuan, Yike and Wang, Jiaqi and He, Conghui and Liu, Ziwei and Chen, Kai and Lin, Dahua},
  booktitle={Computer Vision -- {ECCV} 2024},
  pages={216--233},
  year={2024},
  publisher={Springer}
}

@inproceedings{li2017crowdsourced,
  title={Crowdsourced data management: Overview and challenges},
  author={Li, Guoliang and Zheng, Yudian and Fan, Ju and Wang, Jiannan and Cheng, Reynold},
  booktitle={Proceedings of the 2017 ACM international conference on Management of Data},
  pages={1711--1716},
  year={2017}
}

@article{muller2017retrieval,
  title={Retrieval from and understanding of large-scale multi-modal medical datasets: a review},
  author={M{\"u}ller, Henning and Unay, Devrim},
  journal={IEEE transactions on multimedia},
  volume={19},
  number={9},
  pages={2093--2104},
  year={2017},
  publisher={IEEE}
}

@article{Huang_2024,
   title={A Survey on Hallucination in Large Language Models: Principles, Taxonomy, Challenges, and Open Questions},
   ISSN={1558-2868},
   url={http://dx.doi.org/10.1145/3703155},
   DOI={10.1145/3703155},
   journal={ACM Transactions on Information Systems},
   publisher={Association for Computing Machinery (ACM)},
   author={Huang, Lei and Yu, Weijiang and Ma, Weitao and Zhong, Weihong and Feng, Zhangyin and Wang, Haotian and Chen, Qianglong and Peng, Weihua and Feng, Xiaocheng and Qin, Bing and Liu, Ting},
   year={2024}
}

@inproceedings{
xu2022learning,
title={Learning to Break the Loop: Analyzing and Mitigating Repetitions for Neural Text Generation },
author={Jin Xu and Xiaojiang Liu and Jianhao Yan and Deng Cai and Huayang Li and Jian Li},
booktitle={Advances in Neural Information Processing Systems},
editor={Alice H. Oh and Alekh Agarwal and Danielle Belgrave and Kyunghyun Cho},
year={2022},
url={https://openreview.net/forum?id=sexfswCc7B}
}

@inproceedings{musicgen,
title={Simple and Controllable Music Generation},
author={Jade Copet and Felix Kreuk and Itai Gat and Tal Remez and David Kant and Gabriel Synnaeve and Yossi Adi and Alexandre D{\'e}fossez},
booktitle={Thirty-seventh Conference on Neural Information Processing Systems},
year={2023},
url={https://openreview.net/forum?id=jtiQ26sCJi}
}

@inproceedings{sd3,
      title={Scaling Rectified Flow Transformers for High-Resolution Image Synthesis},
      author={Esser, Patrick and Kulal, Sumith and Blattmann, Andreas and Entezari, Rahim and M{\"u}ller, Jonas and Saini, Harry and Levi, Yam and Lorenz, Dominik and Sauer, Axel and Boesel, Frederic and Podell, Dustin and Dockhorn, Tim and English, Zion and Rombach, Robin},
      booktitle={Proceedings of the 41st International Conference on Machine Learning},
      series={Proceedings of Machine Learning Research},
      volume={235},
      pages={12606--12633},
      year={2024},
      publisher={PMLR}
}

@misc{touvron2023llama2,
      title={Llama 2: Open Foundation and Fine-Tuned Chat Models},
      author={Hugo Touvron and Louis Martin and Kevin Stone and Peter Albert and Amjad Almahairi and Yasmine Babaei and Nikolay Bashlykov and Soumya Batra and Prajjwal Bhargava and Shruti Bhosale and Dan Bikel and Lukas Blecher and Cristian Canton Ferrer and Moya Chen and Guillem Cucurull and David Esiobu and Jude Fernandes and Jeremy Fu and Wenyin Fu and Brian Fuller and Cynthia Gao and Vedanuj Goswami and Naman Goyal and Anthony Hartshorn and Saghar Hosseini and Rui Hou and Hakan Inan and Marcin Kardas and Viktor Kerkez and Madian Khabsa and Isabel Kloumann and Artem Korenev and Punit Singh Koura and Marie-Anne Lachaux and Thibaut Lavril and Jenya Lee and Diana Liskovich and Yinghai Lu and Yuning Mao and Xavier Martinet and Todor Mihaylov and Pushkar Mishra and Igor Molybog and Yixin Nie and Andrew Poulton and Jeremy Reizenstein and Rashi Rungta and Kalyan Saladi and Alan Schelten and Ruan Silva and Eric Michael Smith and Ranjan Subramanian and Xiaoqing Ellen Tan and Binh Tang and Ross Taylor and Adina Williams and Jian Xiang Kuan and Puxin Xu and Zheng Yan and Iliyan Zarov and Yuchen Zhang and Angela Fan and Melanie Kambadur and Sharan Narang and Aurelien Rodriguez and Robert Stojnic and Sergey Edunov and Thomas Scialom},
      year={2023},
      eprint={2307.09288},
      archivePrefix={arXiv},
      primaryClass={cs.CL}
}

@misc{grattafiori2024llama3,
      title={The Llama 3 Herd of Models},
      author={Grattafiori, Aaron and Dubey, Abhimanyu and Jauhri, Abhinav and Pandey, Abhinav and Kadian, Abhishek and Al-Dahle, Ahmad and Letman, Aiesha and Mathur, Akhil and Schelten, Alan and Vaughan, Alex and others},
      year={2024},
      eprint={2407.21783},
      archivePrefix={arXiv},
      primaryClass={cs.AI}
}

@inproceedings{radford2022robustspeechrecognitionlargescale,
      title={Robust Speech Recognition via Large-Scale Weak Supervision},
      author={Radford, Alec and Kim, Jong Wook and Xu, Tao and Brockman, Greg and McLeavey, Christine and Sutskever, Ilya},
      booktitle={Proceedings of the 40th International Conference on Machine Learning},
      series={Proceedings of Machine Learning Research},
      volume={202},
      pages={28492--28518},
      year={2023},
      publisher={PMLR}
}

@misc{verma2024whispergpthybridrepresentationaudio,
      title={Whisper-GPT: A Hybrid Representation Audio Large Language Model}, 
      author={Prateek Verma},
      year={2024},
      eprint={2412.11449},
      archivePrefix={arXiv},
      primaryClass={cs.SD},
      url={https://arxiv.org/abs/2412.11449}, 
}

@misc{du2024lauragptlistenattendunderstand,
      title={LauraGPT: Listen, Attend, Understand, and Regenerate Audio with GPT}, 
      author={Zhihao Du and Jiaming Wang and Qian Chen and Yunfei Chu and Zhifu Gao and Zerui Li and Kai Hu and Xiaohuan Zhou and Jin Xu and Ziyang Ma and Wen Wang and Siqi Zheng and Chang Zhou and Zhijie Yan and Shiliang Zhang},
      year={2023},
      eprint={2310.04673},
      archivePrefix={arXiv},
      primaryClass={cs.SD},
      url={https://arxiv.org/abs/2310.04673}, 
}

@misc{chu2023qwenaudioadvancinguniversalaudio,
      title={Qwen-Audio: Advancing Universal Audio Understanding via Unified Large-Scale Audio-Language Models}, 
      author={Yunfei Chu and Jin Xu and Xiaohuan Zhou and Qian Yang and Shiliang Zhang and Zhijie Yan and Chang Zhou and Jingren Zhou},
      year={2023},
      eprint={2311.07919},
      archivePrefix={arXiv},
      primaryClass={eess.AS},
      url={https://arxiv.org/abs/2311.07919}, 
}

@inproceedings{gong2023jointaudiospeechunderstanding,
      title={Joint Audio and Speech Understanding},
      author={Gong, Yuan and Liu, Alexander H. and Luo, Hongyin and Karlinsky, Leonid and Glass, James},
      booktitle={2023 IEEE Automatic Speech Recognition and Understanding Workshop (ASRU)},
      year={2023},
      publisher={IEEE}
}

@misc{openai2024gpt4ocard,
      title={GPT-4o System Card}, 
      author={OpenAI},
      year={2024},
      eprint={2410.21276},
      archivePrefix={arXiv},
      primaryClass={cs.CL},
      url={https://arxiv.org/abs/2410.21276}, 
}

@inproceedings{soucek2024genhowto,
    title={GenHowTo: Learning to Generate Actions and State Transformations from Instructional Videos},
    author={Sou\v{c}ek, Tom\'{a}\v{s} and Damen, Dima and Wray, Michael and Laptev, Ivan and Sivic, Josef},
    booktitle={Proceedings of the IEEE Conference on Computer Vision and Pattern Recognition (CVPR)},
    month={June},
    year={2024}
}

@misc{ye2023mplugdocowl,
      title={mPLUG-DocOwl: Modularized Multimodal Large Language Model for Document Understanding}, 
      author={Jiabo Ye and Anwen Hu and Haiyang Xu and Qinghao Ye and Ming Yan and Yuhao Dan and Chenlin Zhao and Guohai Xu and Chenliang Li and Junfeng Tian and Qian Qi and Ji Zhang and Fei Huang},
      year={2023},
      eprint={2307.02499},
      archivePrefix={arXiv},
      primaryClass={cs.CL}
}

@inproceedings{li2024fire,
title={{FIRE}: A Dataset for Feedback Integration and Refinement Evaluation of Multimodal Models},
author={Pengxiang Li and Zhi Gao and Bofei Zhang and Tao Yuan and Yuwei Wu and Mehrtash Harandi and Yunde Jia and Song-Chun Zhu and Qing Li},
booktitle={The Thirty-eight Conference on Neural Information Processing Systems Datasets and Benchmarks Track},
year={2024},
url={https://openreview.net/forum?id=DFb1gwnhQS}
}

@inproceedings{yin2023lamm,
    title={{LAMM}: Language-Assisted Multi-Modal Instruction-Tuning Dataset, Framework, and Benchmark},
    author={Yin, Zhenfei and Wang, Jiong and Cao, Jianjian and Shi, Zhelun and Liu, Dingning and Li, Mukai and Huang, Xiaoshui and Wang, Zhiyong and Sheng, Lu and Bai, Lei and Shao, Jing and Ouyang, Wanli},
    booktitle={Advances in Neural Information Processing Systems (NeurIPS) Datasets and Benchmarks Track},
    year={2023}
}

@inproceedings{zhang2024multimodal,
  title={Multimodal Self-Instruct: Synthetic Abstract Image and Visual Reasoning Instruction Using Language Model},
  author={Zhang, Wenqi and Cheng, Zhenglin and He, Yuanyu and Wang, Mengna and Shen, Yongliang and Tan, Zeqi and Hou, Guiyang and He, Mingqian and Ma, Yanna and Lu, Weiming and Zhuang, Yueting},
  booktitle={Proceedings of the 2024 Conference on Empirical Methods in Natural Language Processing},
  pages={19228--19252},
  year={2024},
  publisher={Association for Computational Linguistics}
}

@inproceedings{wang2023vigc,
      title={{VIGC}: Visual Instruction Generation and Correction},
      author={Wang, Bin and Wu, Fan and Han, Xiao and Peng, Jiahui and Zhong, Huaping and Zhang, Pan and Dong, Xiaoyi and Li, Weijia and Li, Wei and Wang, Jiaqi and He, Conghui},
      booktitle={Proceedings of the AAAI Conference on Artificial Intelligence},
      volume={38},
      number={6},
      pages={5309--5317},
      year={2024}
}

@inproceedings{wang2022self,
  title={Self-Instruct: Aligning Language Models with Self-Generated Instructions},
  author={Wang, Yizhong and Kordi, Yeganeh and Mishra, Swaroop and Liu, Alisa and Smith, Noah A. and Khashabi, Daniel and Hajishirzi, Hannaneh},
  booktitle={Proceedings of the 61st Annual Meeting of the Association for Computational Linguistics (Volume 1: Long Papers)},
  pages={13484--13508},
  year={2023},
  publisher={Association for Computational Linguistics}
}

@inproceedings{wu24next,
  title={NExT-GPT: Any-to-Any Multimodal LLM},
  author={Wu, Shengqiong and Fei, Hao and Qu, Leigang and Ji, Wei and Chua, Tat-Seng},
  booktitle={Proceedings of the International Conference on Machine Learning},
  pages = {53366--53397},
  year={2024}
}

@inproceedings{xie2024showo,
  title={Show-o: One Single Transformer to Unify Multimodal Understanding and Generation},
  author={Xie, Jinheng and Mao, Weijia and Bai, Zechen and Zhang, David Junhao and Wang, Weihao and Lin, Kevin Qinghong and Gu, Yuchao and Chen, Zhijie and Yang, Zhenheng and Shou, Mike Zheng},
  booktitle={International Conference on Learning Representations (ICLR)},
  year={2025}
}

@article{ge2024seed,
  title={SEED-X: Multimodal Models with Unified Multi-granularity Comprehension and Generation},
  author={Ge, Yuying and Zhao, Sijie and Zhu, Jinguo and Ge, Yixiao and Yi, Kun and Song, Lin and Li, Chen and Ding, Xiaohan and Shan, Ying},
  journal={arXiv preprint arXiv:2404.14396},
  year={2024}
}

@inproceedings{liu2024visual,
  title={Visual Instruction Tuning},
  author={Liu, Haotian and Li, Chunyuan and Wu, Qingyang and Lee, Yong Jae},
  booktitle={Advances in Neural Information Processing Systems (NeurIPS)},
  volume={36},
  year={2023}
}

@inproceedings{cui2024survey,
  title={A survey on multimodal large language models for autonomous driving},
  author={Cui, Can and Ma, Yunsheng and Cao, Xu and Ye, Wenqian and Zhou, Yang and Liang, Kaizhao and Chen, Jintai and Lu, Juanwu and Yang, Zichong and Liao, Kuei-Da and others},
  booktitle={Proceedings of the IEEE/CVF Winter Conference on Applications of Computer Vision},
  pages={958--979},
  year={2024}
}

@inproceedings{tang2024hawk,
  title={{HAWK}: Learning to Understand Open-World Video Anomalies},
  author={Tang, Jiaqi and Lu, Hao and Wu, Ruizheng and Xu, Xiaogang and Ma, Ke and Fang, Cheng and Guo, Bin and Lu, Jiangbo and Chen, Qifeng and Chen, Ying-Cong},
  booktitle={Advances in Neural Information Processing Systems (NeurIPS)},
  volume={37},
  pages={139751--139785},
  year={2024}
}

@article{li2023videochat,
  title={Videochat: Chat-centric video understanding},
  author={Li, KunChang and He, Yinan and Wang, Yi and Li, Yizhuo and Wang, Wenhai and Luo, Ping and Wang, Yali and Wang, Limin and Qiao, Yu},
  journal={arXiv preprint arXiv:2305.06355},
  year={2023}
}

@article{li2024embodied,
  title={Embodied Intelligence in Mining: Leveraging Multi-modal Large Language Model for Autonomous Driving in Mines},
  author={Li, Luxi and Li, Yuchen and Zhang, Xiaotong and He, Yuhang and Yang, Jianjian and Tian, Bin and Ai, Yunfeng and Li, Lingxi and N{\"u}chter, Andreas and Xuanyuan, Zhe},
  journal={IEEE Transactions on Intelligent Vehicles},
  year={2024},
  publisher={IEEE}
}

@inproceedings{survey1,
author = {Liang, Zijing and Xu, Yanjie and Hong, Yifan and Shang, Penghui and Wang, Qi and Fu, Qiang and Liu, Ke},
title = {A Survey of Multimodel Large Language Models},
year = {2024},
isbn = {9798400716942},
publisher = {Association for Computing Machinery},
address = {New York, NY, USA},
url = {https://doi.org/10.1145/3672758.3672824},
doi = {10.1145/3672758.3672824},
booktitle = {Proceedings of the 3rd International Conference on Computer, Artificial Intelligence and Control Engineering},
pages = {405–409},
numpages = {5},
location = {Xi' an, China},
series = {CAICE '24}
}

@inproceedings{zhang2025oasis,
  title={{Oasis}: One Image is All You Need for Multimodal Instruction Data Synthesis},
  author={Zhang, Letian and Cui, Quan and Zhao, Bingchen and Yang, Cheng},
  booktitle={Proceedings of the IEEE/CVF International Conference on Computer Vision ({ICCV})},
  pages={3542--3551},
  year={2025}
}

@article{hansen2025instructify,
  title={{Instructify}: Demystifying Metadata to Visual Instruction Tuning Data Conversion},
  author={Hansen, Jacob and Lin, Wei and Kang, Junmo and Mirza, Muhammad Jehanzeb and Luo, Hongyin and Feris, Rogerio and Ritter, Alan and Glass, James and Karlinsky, Leonid},
  journal={arXiv preprint arXiv:2505.18115},
  year={2025}
}

@inproceedings{ding2025mmifengine,
  title={{MM-IFEngine}: Towards Multimodal Instruction Following},
  author={Ding, Shengyuan and Wu, Shenxi and Zhao, Xiangyu and Zang, Yuhang and Duan, Haodong and Dong, Xiaoyi and Zhang, Pan and Cao, Yuhang and Lin, Dahua and Wang, Jiaqi},
  booktitle={Proceedings of the IEEE/CVF International Conference on Computer Vision ({ICCV})},
  pages={1099--1109},
  year={2025}
}

@inproceedings{feng2026intersyn,
  title={A High Quality Dataset and Reliable Evaluation for Interleaved Image-Text Generation},
  author={Feng, Yukang and Sun, Jianwen and Li, Chuanhao and Li, Zizhen and Ai, Jiaxin and Zhang, Fanrui and Chang, Yifan and Zhou, Sizhuo and Zhang, Shenglin and Dai, Yu and Zhang, Kaipeng},
  booktitle={International Conference on Learning Representations ({ICLR})},
  year={2026}
}

\clearpage
\newpage

\appendix

\section*{Overview of the Appendix}
\label{app:overview}

This appendix is organized as follows:

\begin{itemize}
  \item \textbf{Dataset construction} (Appendix~\ref{app:dataset}): the topic pool that seeds generation, the external multimodal toolkit, dataset diversity, bias-mitigation procedures, and modality extensibility.
  \item \textbf{Methodology} (Appendix~\ref{app:method}): a worked example of the instruction-flow correction step and an analysis of why \name\ improves over a vanilla GPT-4o pipeline.
  \item \textbf{Experiments} (Appendix~\ref{app:exp}): training and inference recipes, evaluation prompts, a user study, out-of-distribution results, and ablations on multimodal balance and user-conditioned generation.
  \item \textbf{Additional analyses}: a scope comparison with recent instruction-data pipelines (Appendix~\ref{app:recent-baselines}).
  \item \textbf{Future work} (Appendix~\ref{app:future}) and additional qualitative examples (Figures~\ref{a0}--\ref{a3}).
\end{itemize}

\section{Dataset Construction}
\label{app:dataset}

\begin{table*}
\centering
\small
\caption{The 90-domain topic pool used to seed \namedataset\ generation. At sampling time two domains are intersected (Eq.~\ref{eq:topic-com}) to form a composite theme.}
\label{tab:topic_pool}
\setlength{\extrarowheight}{1pt}
\rowcolors{1}{gray!10}{white}
\begin{tabularx}{\linewidth}{*{6}{>{\centering\arraybackslash}X}}
\toprule
e-commerce & academic discussions & lifestyle & sports & mathematics & business \\
technology & travels & health & entertainment & art & history \\
cooking & parenting & fashion & finance & politics & literature \\
gardening & astronomy & music & education & computer science & programming \\
design & research & internship & food & beauty & entrepreneurship \\
startups & algorithm & bugs & scholar & IT & vision \\
supervision & classroom & assignment & game & psychology & social media \\
film & theater & architecture & photography & environment & agriculture \\
real estate & marketing & economics & philosophy & law & medicine \\
engineering & physics & biology & robotics & cybersecurity & artificial intelligence \\
family & space exploration & climate change & urban planning & transportation & hospitality \\
tourism & fitness & relationships & holidays & charity & DIY projects \\
survival skills & life & cartoon & pets & wildlife & marine biology \\
genetics & neuroscience & anthropology & sociology & linguistics & culture \\
mythology & archaeology & forensic science & criminology & military & diplomacy \\
\bottomrule
\end{tabularx}
\end{table*}

\begin{table*}
\centering
\small
\caption{External tools used by \name\ during multimodal realization. Each tool is responsible for one modality, allowing the pipeline to be extended modularly.}
\label{tab:toolkit}
\setlength{\tabcolsep}{6pt}
\setlength{\extrarowheight}{1pt}
\begin{tabularx}{\linewidth}{l|l|l|X}
\toprule
\textbf{Tool} & Modality & Source & Role in the pipeline \\
\midrule
\textbf{GPT-4o}~\cite{openai2024gpt4ocard} & \texttt{<language>} & OpenAI API & Orchestrator and language backbone; expands events, drafts dialogue, and validates outputs. \\
\textbf{Google Search} & \texttt{<link>} & \url{https://www.google.com} & Provides up-to-date factual snippets and reference URLs. \\
\textbf{Google Maps} & \texttt{<map>} & \url{https://www.google.com/maps} & Supplies geographical references and routing information. \\
\textbf{DALL$\cdot$E\,3} & \texttt{<image>} & \url{https://openai.com/index/dall-e-3/} & Synthesizes photo-realistic and stylized images from text prompts. \\
\textbf{MusicGen}~\cite{musicgen} & \texttt{<music>} & Open-source model & Generates instrumental clips from textual descriptions. \\
\textbf{Apple Emoji Archive} & \texttt{<emoji>} & \url{https://www.apple.com/} & Provides culturally familiar emoji symbols for affective cues. \\
\bottomrule
\end{tabularx}
\end{table*}

\subsection{Topic Pool}
\label{app:topic}

We curate a \textbf{topic pool} of 90 domains spanning everyday life, science and engineering, business, the arts, and the humanities (Table~\ref{tab:topic_pool}). The breadth of this pool is the primary driver of \namedataset's content diversity, as it determines the universe of scenarios from which subsequent events are drawn.

For each instruction sample we draw two topics $Q_1, Q_2$ uniformly at random and merge them into a composite theme,
\begin{equation}
\label{eq:topic-com}
  \langle\textit{topic}\rangle \;=\; Q_1 \cap Q_2.
\end{equation}
Intersecting unrelated domains (for instance, \emph{cooking} $\cap$ \emph{chemistry} or \emph{music} $\cap$ \emph{marketing}) introduces controlled stochasticity into the pipeline and yields interdisciplinary prompts that are difficult to obtain from any single-domain source.

\subsection{Multimodal Toolkit}
\label{app:toolkit}

To realize the any-to-any generation interface described in the main paper, \name\ orchestrates a suite of off-the-shelf tools, each responsible for a single modality. Table~\ref{tab:toolkit} summarizes every component, its target modality, its source, and the specific role it plays in the pipeline. We deliberately favor publicly available APIs and open-source models so that the pipeline can be reproduced and extended by the community.

\subsection{Data Diversity}
\label{app:diversity}

Table~\ref{datacompared} compares \namedataset\ with widely-used multimodal instruction datasets along three orthogonal axes: (i) the number of dialogue rounds per sample, (ii) the count of distinct input/output modalities, and (iii) the overall corpus size. \namedataset\ supports nine modalities on both the input and the output side, the largest modality set in this comparison, and provides on average $17.5$ rounds per dialogue (vs.\ at most $4.8$ for prior work).

\begin{table*}
  \centering
  \caption{Comparison with current multimodal instruction datasets. \emph{Num} is the number of distinct modality types; \emph{Class} lists the modalities themselves. Legend: L:\texttt{<language>}; I:\texttt{<image>}; V:\texttt{<video>}; 3D:\texttt{<point cloud>}; LK:\texttt{<link>}; A:\texttt{<audio>}; M:\texttt{<music>}; E:\texttt{<emoji>}; C:\texttt{<code>}; MP:\texttt{<map>}; MT:\texttt{<math>}; Q:\texttt{<QR code>}.}
  \label{datacompared}
  \Large
  \resizebox{\linewidth}{!}{%
  \begin{tabular}{l|cc|cc|c|c}
  \toprule
  \multirow{2}[2]{*}{\textbf{Dataset}} & \multicolumn{2}{c|}{Input Modality} & \multicolumn{2}{c|}{Output Modality} & \multirow{2}[2]{*}{Avg. \#Rounds} & \multirow{2}[2]{*}{\#Instruction} \\
  \cmidrule(lr){2-3}\cmidrule(lr){4-5}
   & Num & Class & Num & Class & & \\
  \midrule
  \textbf{LLaVA}~\cite{liu2024visual}            & 2 & L/I     & 1 & L       & $\approx$ 1.0 & \phantom{00}5K  \\
  \textbf{VideoChat}~\cite{li2023videochat}      & 2 & L/V     & 1 & L       & $\approx$ 1.8 & \phantom{0}11K  \\
  \textbf{FIRE}~\cite{li2024fire}                & 2 & L/I     & 1 & L       & $\approx$ 1.0 & 100K            \\
  \textbf{LAMM}~\cite{yin2023lamm}               & 3 & L/I/3D  & 1 & L       & $\approx$ 3.3 & 196K            \\
  \textbf{mPLUG-DocOwl}~\cite{ye2023mplugdocowl} & 4 & L/I/T/L & 1 & L       & -             & -               \\
  \midrule
  \textbf{GenHowTo}~\cite{soucek2024genhowto}    & 2 & L/I     & 2 & L/I     & -             & -               \\
  \textbf{MosIT}~\cite{wu24next}                 & 4 & L/I/A/V & 4 & L/I/A/V & $\approx$ 4.8 & \phantom{00}5K  \\
  \midrule
  \textbf{Ours (\namedataset)} & \textcolor{red}{9} & L/I/M/E/C/MP/LK/MT/Q & \textcolor{red}{9} & L/I/M/E/C/MP/LK/MT/Q & $\approx$ \textcolor{red}{17.5} & \textcolor{red}{20K} \\
  \bottomrule
  \end{tabular}}
\end{table*}

\subsection{Human Oversight for Bias Mitigation}
\label{app:bias}

To reduce the bias inherited from any single LLM, we insert lightweight human oversight at two carefully chosen points in the pipeline.

\paragraph{Topic curation.} The 90 seed topics are hand-selected to balance category coverage and cultural breadth, preventing the corpus from skewing toward a narrow set of domains.

\paragraph{Event filtering.} After event expansion, annotators discard implausible events and remove near-duplicates. Cleaner events produce better downstream prompts and, importantly, mitigate the propagation of LLM-specific stylistic biases into the final instructions.

\subsection{Modality Extensibility}
\label{app:extension}

The pipeline is designed to absorb new modalities without architectural changes.

\paragraph{Stage~3 (raw-instruction generation).} The planner emits modality-agnostic slots that are filled at realization time, so adding a modality beyond the current nine does not require changing the planner.

\paragraph{Stage~4 (multimodal realization).} New APIs (such as Text-to-Video or Text-to-3D models) can be plugged in via a thin adapter that maps modality slots to API calls, with no change to upstream components.

\section{Methodology Details}
\label{app:method}

\subsection{Instruction-Flow Correction in Practice}
\label{app:correction}

Correction is performed \emph{before} multimodal realization. At this stage every multimodal slot is still represented by its textual prompt rather than the rendered media, so the corrector can reason about content, dependencies, and redundancy purely at the language level. A representative snippet is shown below.

\begin{promptbox}{Example: textualized multimodal slot during correction}
\textbf{Q}: \ldots \\[2pt]
\textbf{A}: This further stimulates economic growth by expanding markets and reducing food scarcity issues. \\[2pt]
\texttt{\textbf{<|image|>}}\, \emph{Prompt: farmland used for marketing campaigns.}\, \texttt{\textbf{<|image\_end|>}}
\end{promptbox}

Because the textual prompt is already available at this stage, the corrector can detect redundancy or topical drift without inspecting the final media, which keeps correction fast and cheap. A tighter feedback loop that conditions the corrector on the rendered media itself is left to future work.

\subsection{Why \name\ Outperforms a Vanilla Pipeline}
\label{app:vs-gpt4o}

\paragraph{Richer multimodal toolkit.} \name\ couples GPT-4o with an external toolkit (DALL$\cdot$E for images, MusicGen for audio, search APIs for facts), extending the effective modality range far beyond what a single language model can produce natively.

\paragraph{Pipeline design.} In Table~\ref{tab:ood_performance}, Self-Instruct runs on a GPT-4 backbone while \name\ runs on LLaMA-3 7B, yet \name\ scores higher on all four GPT-guided metrics. Under this comparison the gain is attributable to event expansion and instruction-flow correction rather than to backbone capacity.

\FloatBarrier
\section{Experimental Details}
\label{app:exp}

\subsection{Training and Inference}
\label{app:train}

\paragraph{Training of $\mathbf{DEG}$.} The Diverse Event Generator (DEG) is fine-tuned on top of LLaMA-3 using $\langle\textit{topic}\rangle \rightarrow \langle\textit{event}\rangle$ pairs harvested from \namedataset. Fine-tuning takes roughly 40 hours on a single NVIDIA A100 (80\,GB).

\paragraph{Training of $\mathbf{MIG}$.} The Multimodal Instruction Generator (MIG) is fine-tuned on four NVIDIA A100 (80\,GB) GPUs for roughly 40 hours.

\subsection{Evaluation Prompt and Metrics}
\label{app:metrics}

We define four GPT-4o-judged metrics (\emph{Reasonableness}, \emph{Clarity}, \emph{Detail}, and \emph{Relevance}) and query the judge with the unified prompt shown below.

\begin{promptbox}{Unified GPT-4o evaluation prompt}
Evaluate the \texttt{\{metric\}} of the multimodal instruction \textbf{\{generated\}}, compared to the ground truth \textbf{\{reference\}}. Return a single score in $[0, 1]$.\\[2pt]
Example output: \verb|{"score": 0.675}|.\\[2pt]
Definition of \texttt{\{metric\}}: \textbf{\{explanation\}}.
\end{promptbox}

The four metrics are operationalized as follows:

\begin{itemize}
  \item \textbf{Reasonableness}: logical consistency and factual accuracy with respect to the reference.
  \item \textbf{Clarity}: readability, well-formedness, and unambiguity of the generated text.
  \item \textbf{Detail}: richness of information conveyed in each turn.
  \item \textbf{Relevance}: topical adherence to the input context across the full dialogue.
\end{itemize}

The evaluator is asked to score each metric independently so that absolute and relative comparisons can be made across the four dimensions.

\subsection{User Study}
\label{app:user-study}

To complement the automatic GPT-4o scores, we run a forced-choice user study. We recruit 15 evaluators from a mix of academic and industrial backgrounds, all with prior experience with large language models. Each evaluator inspects 50 samples and, for every one of the four evaluation dimensions, selects which method among Self-Instruct~\cite{wang2022self}, VIGC~\cite{wang2023vigc}, and \name\ produced the best output. The results, reported in Table~\ref{tab:user_study_results}, show that \name\ is preferred by a substantial margin on every dimension.

\begin{table}[H]
\centering
\caption{User study: percentage of samples (\%) on which each method is selected as the best by 15 human evaluators across 50 samples (per-dimension forced choice). \textcolor{red}{Red} indicates the best performance.}
\label{tab:user_study_results}
\resizebox{\linewidth}{!}{%
\begin{tabular}{l|cccc}
\toprule
\textbf{Method} & Reason. $(\mathbf{\uparrow})$ & Clarity $(\mathbf{\uparrow})$ & Detail $(\mathbf{\uparrow})$ & Relev. $(\mathbf{\uparrow})$ \\
\midrule
\textbf{Self-Instruct}~\cite{wang2022self} & 18.5 & \phantom{0}9.7 & \phantom{0}9.8 & \phantom{0}8.5 \\
\textbf{VIGC}~\cite{wang2023vigc}          & 10.3 & \phantom{0}5.3 & 14.4 & 10.9 \\
\midrule
\textbf{Ours (\name)} & \textcolor{red}{71.2} & \textcolor{red}{85.0} & \textcolor{red}{75.8} & \textcolor{red}{80.6} \\
\bottomrule
\end{tabular}}
\end{table}

\begin{table*}[!t]
  \centering
  \caption{Quality of generated instructions on the out-of-distribution (OOD) set. \textbf{Bold} marks the best score in each column. \emph{\#Others} aggregates the emoji, code, math, map, link, and QR-code modalities.}
  \label{tab:ood_performance}
  \Large
  \resizebox{\linewidth}{!}{%
  \begin{tabular}{l|l|cccc|ccc|c}
  \toprule
  \multirow{2}[2]{*}{\textbf{Method}} & \multirow{2}[2]{*}{Backbones}
    & \multicolumn{4}{c|}{GPT-Guided Metrics $(\mathbf{\uparrow})$}
    & \multicolumn{3}{c|}{Avg. \#Multimodalities$(\mathbf{\uparrow})$}
    & \multirow{2}[2]{*}{Avg. \#Rounds} \\
  \cmidrule{3-9}
  & & Reasonableness & Clarity & Detail & Relevance & \#Image & \#Music & \#Others & \\
  \midrule
  \textbf{Self-Instruct}~\cite{wang2022self} & GPT-4~\cite{openai2024gpt4ocard}   & 0.735 & 0.475 & 0.608 & 0.580 & - & - & - & $\approx$ 3 \\
  \textbf{VIGC}~\cite{wang2023vigc}          & Vicuna 7B~\cite{chiang2023vicuna}  & 0.714 & 0.507 & 0.732 & 0.737 & - & - & - & $\approx$ 1 \\
  \midrule
  \textbf{Ours (\name)} & LLaMA-3 7B~\cite{grattafiori2024llama3} & \textcolor{red}{0.752} & \textcolor{red}{0.701} & \textcolor{red}{0.808} & \textcolor{red}{0.845} & \textcolor{red}{1.04} & \textcolor{red}{0.27} & \textcolor{red}{3.82} & $\approx$ \textcolor{red}{10} \\
  \bottomrule
  \end{tabular}}
\end{table*}

\subsection{Out-of-Distribution Performance}
\label{app:ood}

To probe generalization, we construct an out-of-distribution (OOD) test set of 100 events drawn from domains outside the seed topic pool (collected via web scraping followed by manual screening). Table~\ref{tab:ood_performance} shows that \name\ remains the strongest method on every metric while simultaneously producing visibly richer multimodal content (more images, music, and other modalities per dialogue) and longer interactions.

\subsection{Multimodal Distribution across Dialog Position}
\label{app:balance}

A common failure mode of existing pipelines is concentrating multimodal content in the opening turns of a dialogue. Table~\ref{tab:multi_modal_distribution} shows that \namedataset\ instead distributes multimodal content nearly uniformly across the entire dialogue (the five normalized buckets each carry roughly 20\,\% of the multimodal mass). The qualitative examples in Figures~\ref{a1}--\ref{a3} corroborate this: modalities appear consistently in early, middle, and late rounds.

\begin{table}[H]
\centering
\scriptsize
\caption{Fraction of multimodal content (\%) as a function of the normalized position within a dialogue. Position is binned into five equal-width buckets.}
\label{tab:multi_modal_distribution}
\resizebox{\linewidth}{!}{%
\begin{tabular}{c|c}
\toprule
\textbf{Position in dialogue} & Multimodal frequency (\%) \\
\midrule
\textbf{\phantom{0}0\,--\,20\,\%} & 20.5 \\
\textbf{20\,--\,40\,\%}           & 19.4 \\
\textbf{40\,--\,60\,\%}           & 19.5 \\
\textbf{60\,--\,80\,\%}           & 21.3 \\
\textbf{80\,--\,100\,\%}          & 19.3 \\
\bottomrule
\end{tabular}}
\end{table}

\subsection{Adaptation to User-Specified Modalities}
\label{app:adapt}

When the user steers the pipeline toward a particular modality, the generated content shifts accordingly (Table~\ref{tab:modality_adaptation}). For example, an image-focused input yields roughly $15\times$ more image content than the general setting, while a music-focused input boosts the music count by a similar factor. The orchestration layer therefore responds to the requested modality rather than defaulting to a fixed mix.

\begin{table}[h]
\centering
\scriptsize
\caption{Average count of each modality per dialogue when the user specifies a desired modality at input time. \textcolor{red}{Red} indicates the targeted modality in each row.}
\label{tab:modality_adaptation}
\resizebox{\linewidth}{!}{%
\begin{tabular}{l|ccc}
\toprule
\textbf{Input type} & \#Image & \#Music & \#Others \\
\midrule
\textbf{General}       & \phantom{0}1.92 & \phantom{0}0.74 & 7.37 \\
\textbf{Music-related} & \phantom{0}1.10 & \textcolor{red}{12.30} & 2.10 \\
\textbf{Image-related} & \textcolor{red}{14.90} & \phantom{0}0.10 & 5.20 \\
\bottomrule
\end{tabular}}
\end{table}

\subsection{Comparison with Recent Instruction-Data Pipelines}
\label{app:recent-baselines}

The two baselines in Table~\ref{qualtable} (Self-Instruct and VIGC) are chosen because they represent the two dominant styles of universal instruction-data generation: single-modality seed-based expansion and vision-conditioned instruction generation. For completeness, we position \name\ against several recent instruction-data pipelines that appeared alongside or after this work.

\begin{itemize}
  \item \textbf{Oasis}~\cite{zhang2025oasis} generates language-side instruction--response pairs from a single conditioning image; assistant-side output is text-only.
  \item \textbf{Instructify}~\cite{hansen2025instructify} iteratively generates multi-turn visual instruction conversations by aggregating over $40$ source datasets; assistant-side output is again text-only.
  \item \textbf{MM-IFEngine}~\cite{ding2025mmifengine} produces constraint-rich image--instruction--answer triples for SFT and DPO; assistant-side output is text-only.
  \item \textbf{InterSyn}~\cite{feng2026intersyn} synthesizes single- and multi-turn interleaved image--text dialogues, with assistant-side output covering language and image.
\end{itemize}

These pipelines target different source data, tasks, and evaluation protocols, so their paper-reported scores cannot be placed in a common ``quality'' column with those in Table~\ref{qualtable} without rerunning every method under an identical protocol. We therefore treat them as scope comparisons rather than head-to-head numeric competitors. The distinguishing feature of \name\ relative to these pipelines is the combination of \emph{symmetric} nine-modality input/output coverage (rather than a language-only or language+image assistant side) and long multi-round interactions (17.5 rounds on average). Among the recent works listed, InterSyn~\cite{feng2026intersyn} is closest in spirit but still limits assistant-side output to language and image.

\subsection{Additional Qualitative Results}
\label{app:qualitative}

\begin{figure*}[h]
  \centering
  \includegraphics[width=\linewidth]{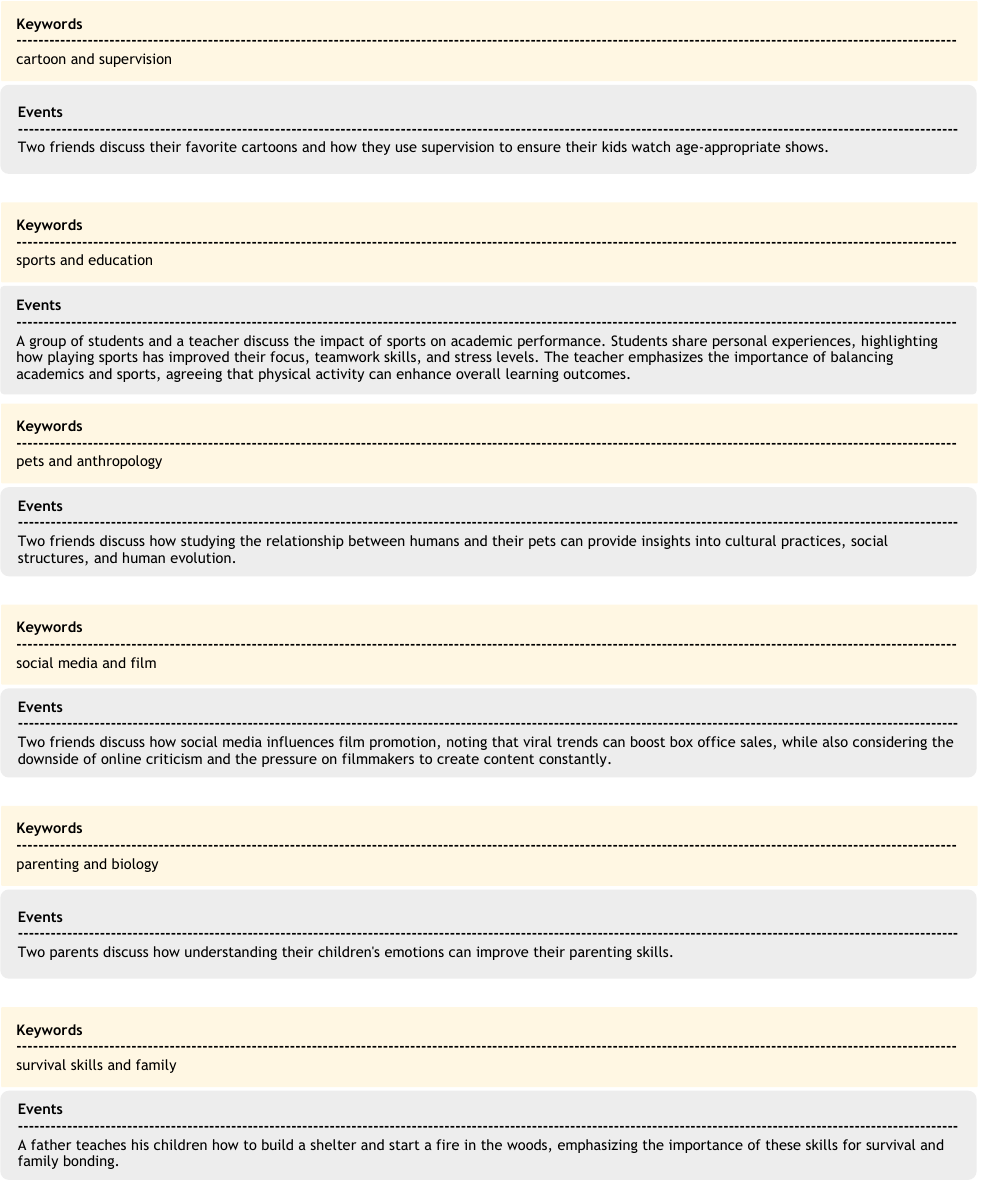}
  \caption{Diverse event generation by \textbf{DEG}. For the same keyword pair, three independently sampled events differ substantially in content, scenario, and intended modality, illustrating the diversity that drives downstream instruction variety.}
  \label{a0}
\end{figure*}

Figure~\ref{a0} visualizes the output diversity of \textbf{DEG}: three independently sampled events for the same keyword pair differ substantially in content, tone, and target modality.

\begin{figure*}[h]
  \centering
  \includegraphics[width=\linewidth]{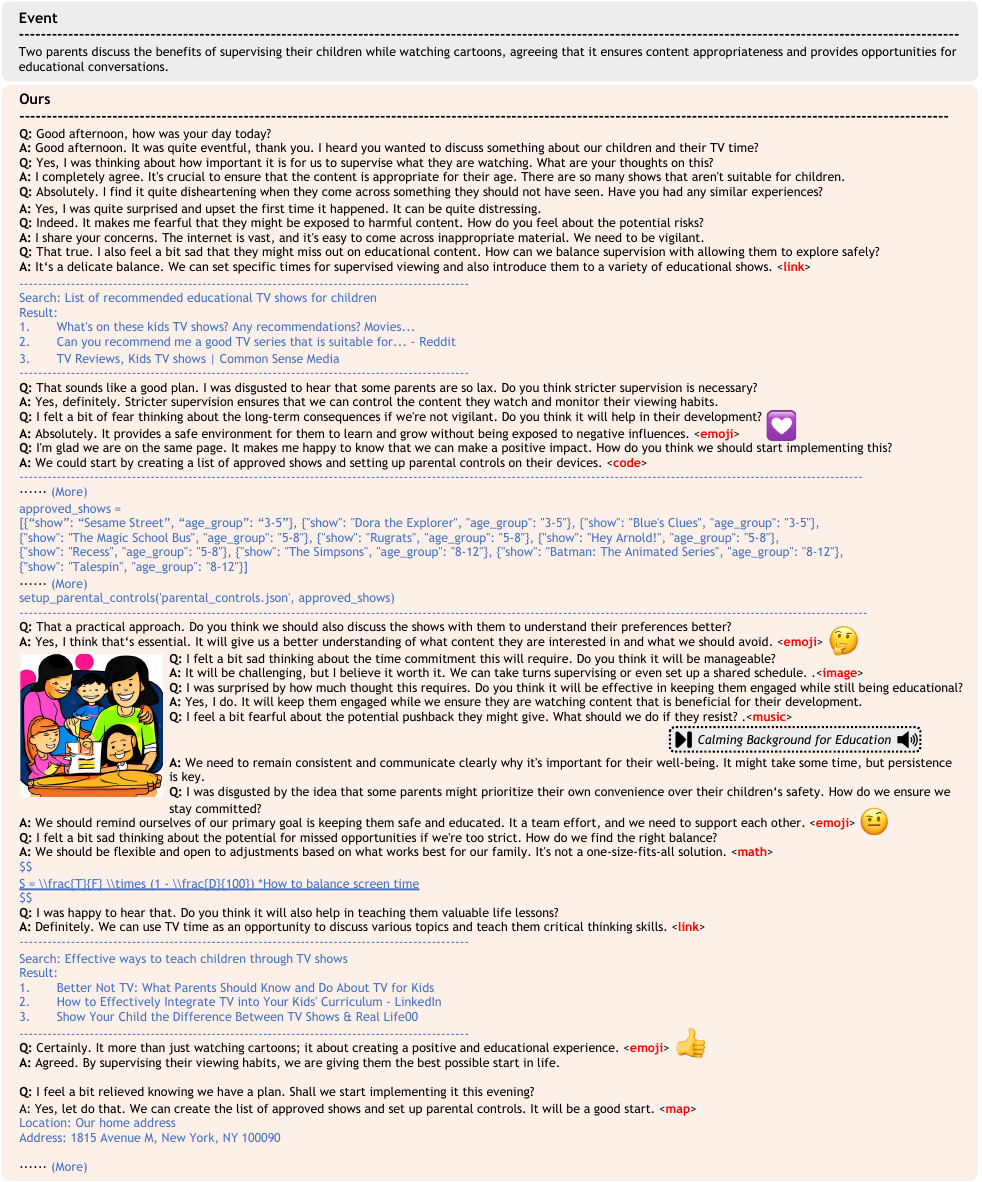}
  \caption{End-to-end qualitative example (I): a multi-round, multimodal instruction produced by \name. Image, music, and text are interleaved into a coherent narrative.}
  \label{a1}
\end{figure*}

\begin{figure*}[h]
  \centering
  \includegraphics[width=\linewidth]{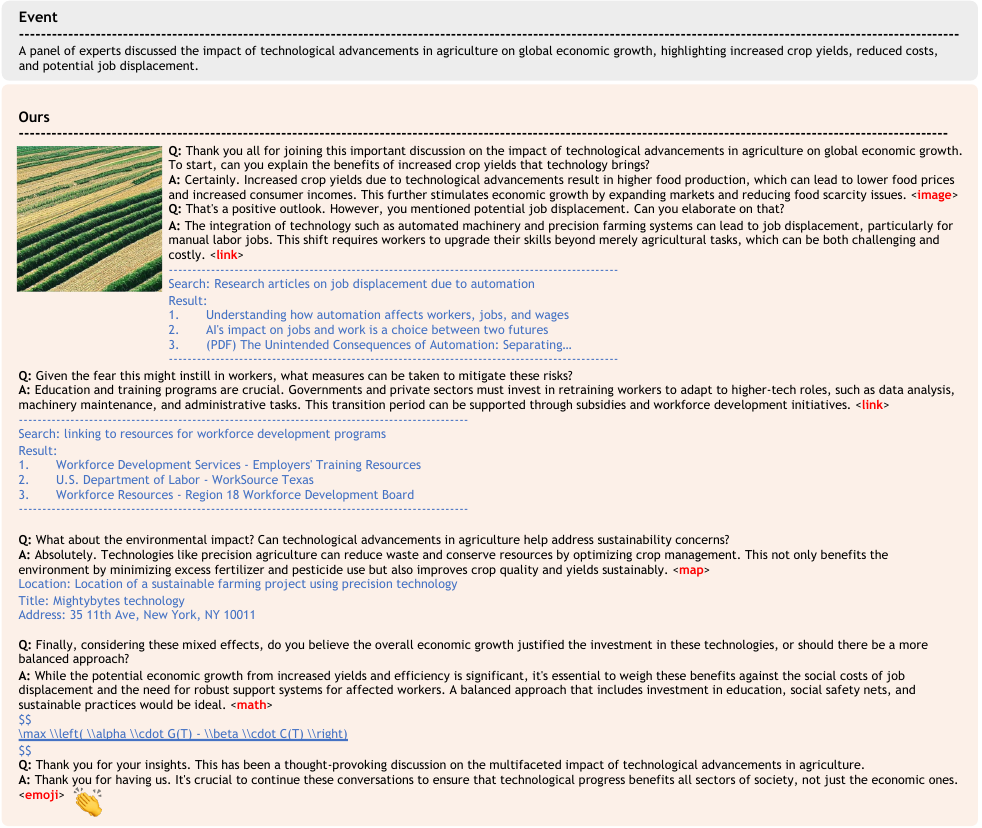}
  \caption{End-to-end qualitative example (II): a different topic pair, showing variation in dialogue length and modality mix while preserving topical coherence.}
  \label{a2}
\end{figure*}

\begin{figure*}[h]
  \centering
  \includegraphics[width=\linewidth]{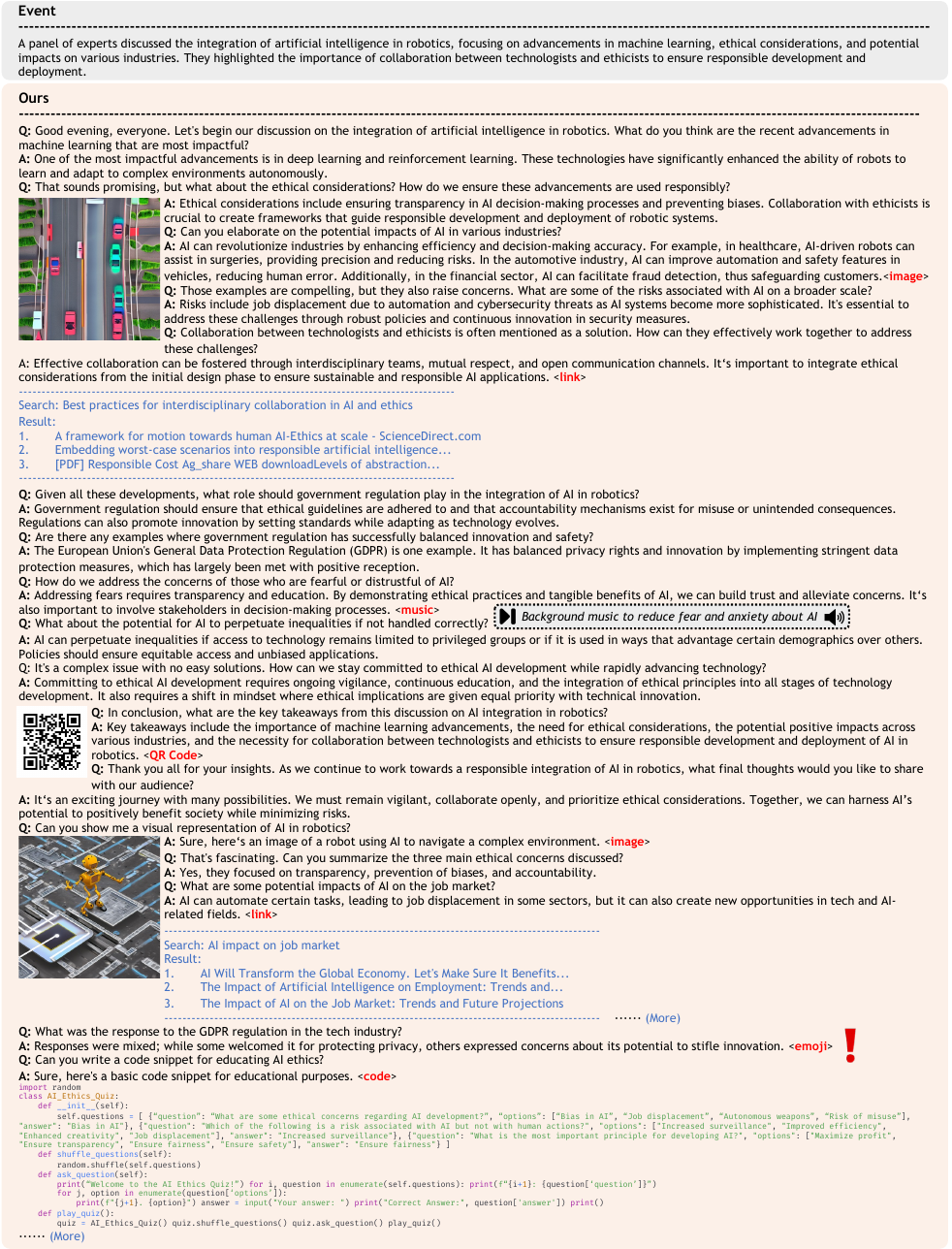}
  \caption{End-to-end qualitative example (III): a long-horizon dialogue that interleaves text, image, code, and emoji content, illustrating \name's behavior on extended interactions.}
  \label{a3}
\end{figure*}

Figures~\ref{a1}--\ref{a3} present three end-to-end examples produced by \textbf{MIG}. Each example interleaves several modalities (image, music, code, emoji) across a multi-round dialogue, illustrating both the modality breadth and the long-horizon coherence achieved by \name.

\FloatBarrier
\section{Future Work}
\label{app:future}

We view \name\ as a foundation for the automated production of multimodal training data, and identify three concrete directions for follow-up work.

\paragraph{Scaling backbones.} Replace the LLaMA-3 backbone of \textbf{MIG} with more recent and larger open-source models to characterize how multimodal instruction quality scales with backbone capacity.

\paragraph{Expanding modalities.} Plug in tools for video, 3D, motion, and tabular generation so that \namedataset\ can extend beyond the current nine modalities, increasing applicability to embodied and scientific domains.

\paragraph{User-conditioned generation.} Tighten the control loop so that \name\ behaves as an interactive data factory: an end user provides their downstream specification, and \name\ produces purpose-built data aligned with that deployment scenario.

\newpage

\newpage

\newpage

\newpage

\end{document}